\documentclass[runningheads]{llncs}

\usepackage{eccv}

\usepackage{eccvabbrv}

\usepackage{graphicx}
\usepackage{booktabs}

\usepackage[accsupp]{axessibility}  %

\usepackage{multirow}
\usepackage{float,wrapfig}
\usepackage{placeins}  %
\usepackage{listings}
\definecolor{eccvblue}{rgb}{0.12,0.49,0.85}
\usepackage[breaklinks,colorlinks,citecolor=eccvblue]{hyperref}
\usepackage{orcidlink}

\definecolor{amethyst}{rgb}{0.6, 0.4, 0.8}
\definecolor{darkpastelgreen}{rgb}{0.01, 0.75, 0.24}
\definecolor{amber}{rgb}{1.0, 0.75, 0.0}
\definecolor{cadmiumorange}{rgb}{0.93, 0.53, 0.18}
\definecolor{lawngreen}{rgb}{0.49, 0.99, 0.0}
\definecolor{limegreen}{rgb}{0.2, 0.8, 0.2}
\definecolor{neongreen}{rgb}{0.22, 0.88, 0.08}
\definecolor{amethyst}{rgb}{0.6, 0.4, 0.8}
\definecolor{darkpastelgreen}{rgb}{0.01, 0.75, 0.24}
\definecolor{greenbest}{RGB}{88,137,15}
\definecolor{redworst}{RGB}{137,15,27}
\definecolor{royalazure}{rgb}{0.25, 0.41, 0.88}

\usepackage{comment}
\newif\ifshowdraft
\showdrafttrue  %

\ifshowdraft
  
\else
  \excludecomment{draft}
\fi

\begin{document}

\title{Global Pose Control for Generative View Synthesis in Normalized Object Coordinate Space} 

\titlerunning{Generative View Synthesis in Normalized Object Coordinate Space}

\author{Zhibing Li\inst{1}\thanks{Work done while Zhibing Li was intern at Amazon.}\orcidlink{0009-0002-4528-5495} \and
Amogh Gupta\inst{2}\orcidlink{0009-0003-0549-8784} \and
Behnoosh Parsa\inst{2}\orcidlink{0000-0003-4788-271X} \and
Dan Casas\inst{2}\orcidlink{0000-0002-3664-089X}}

\authorrunning{Z.~Li et al.}

\institute{The Chinese University of Hong Kong, China \and
Amazon, USA
\\[2pt]
\href{https://lizb6626.github.io/GlobalNVS/}{\texttt{https://lizb6626.github.io/GlobalNVS/}}
}

\maketitle

\begin{abstract}
  Novel View Synthesis (NVS) enables the generation of unseen views of a scene from a single or multiple images, allowing users to freely explore an object from any viewpoint. Despite the recent impressive qualitative improvements of generative models for this task, existing methods struggle to provide global and intuitive control of target viewpoints because they either use input-relative camera poses or are limited to generating sparse global views. This lack of global pose control severely limits the number of downstream tasks potentially enabled by NVS. 
  To address this limitation, we propose a novel approach for precise camera control in a customizable Normalized Object Coordinate Space (NOCS), requiring single or few unposed images. Our method operates solely on the absolute camera pose of the target view in NOCS, eliminating the need for a relative world frame or camera poses of the input images. 
  Unlike previous methods that treat NVS as a standalone generation task, we formulate it as an image editing problem and build upon state-of-the-art editing models to leverage their superior generalization capability. Camera information is injected as dedicated camera tokens via an in-context multi-modal conditioning strategy. To alleviate the inherent ambiguity of NOCS, we incorporate text descriptions that explicitly define the object's canonical coordinate frame, which also enhances generalization to unseen object categories. Furthermore, we curate a high-quality dataset with consistently aligned orientations and corresponding NOCS text definitions. 
  Extensive experiments demonstrate that our method robustly generates novel views with accurate and consistent orientations from arbitrary unposed images across diverse categories, achieving state-of-the-art image quality and fidelity.
\end{abstract}
\section{Introduction}
\label{sec:intro}
Recent advances in generative diffusion models have made novel view synthesis (NVS) tasks from single or multiple images both accessible and effective~\cite{watson2023novel,shi2024mvdream,voleti2024sv3d,zhou2025stable}. 
This progress in unseen image synthesis is enabling rapid construction of models for 3D reconstruction~\cite{long2024wonder3d,liu2024syncdreamer}, video manipulation~\cite{ren2025gen3c,yu2025trajectory,zhang2025recapture,bahmani2025vdd,wang2024motionctrl}, content creation, robotic simulation~\cite{shi2025nvspolicyadaptivenovelviewsynthesis}, and beyond.

The fundamental ---and arguably most desired--- feature in NVS models is their ability to provide a \textit{global} and \textit{intuitive} control of the target viewpoint, however, existing models either use input-relative viewpoint control \cite{liu2023zero1to3,Szymanowicz2023ViewsetD,xu2024grm,tseng2023consistent,zhou2025stable,liu2024syncdreamer,kong2024eschernet} or are limited to generating sparse global views~\cite{li2025droplet3d}.
This prevents NVS models from being used for practical applications at scale, because the user cannot automatically assume a standard \textit{canonical} coordinate frame with respect to which synthesized images are aligned to.  
Image-to-3D models \cite{hunyuan3d2025hunyuan3d, xiang2024structured, long2024wonder3d,liu2023one2345,liu2023one2345++} offer an attractive alternative since the output 3D mesh can be rendered from an arbitrary viewpoint, however they often neglect the mesh orientation alignment altogether, mostly due to inconsistencies in large-scale 3D datasets \cite{Deitke2022ObjaverseAU,xiang2024structured}, resulting in models that lack standard canonical orientations.
Consequently, output 3D meshes cannot be used for tasks that require object-centric global viewpoint control. 

Recent methods \cite{wang2025orient,huang2025cupid} have focused on orientation alignment by fine-tuning image-to-3D models using carefully aligned 3D training sets. 
While they alleviate the alignment issue to some extent, they remain fundamentally constrained by the texture quality of 3D generation, which still falls far behind that of state-of-the-art 2D image synthesis, resulting in outputs that lack fine appearance details and high-fidelity visual quality.

\begin{figure}[t]
    \centering
    
    \includegraphics[trim={55 0 55 0},clip,width=1\linewidth]{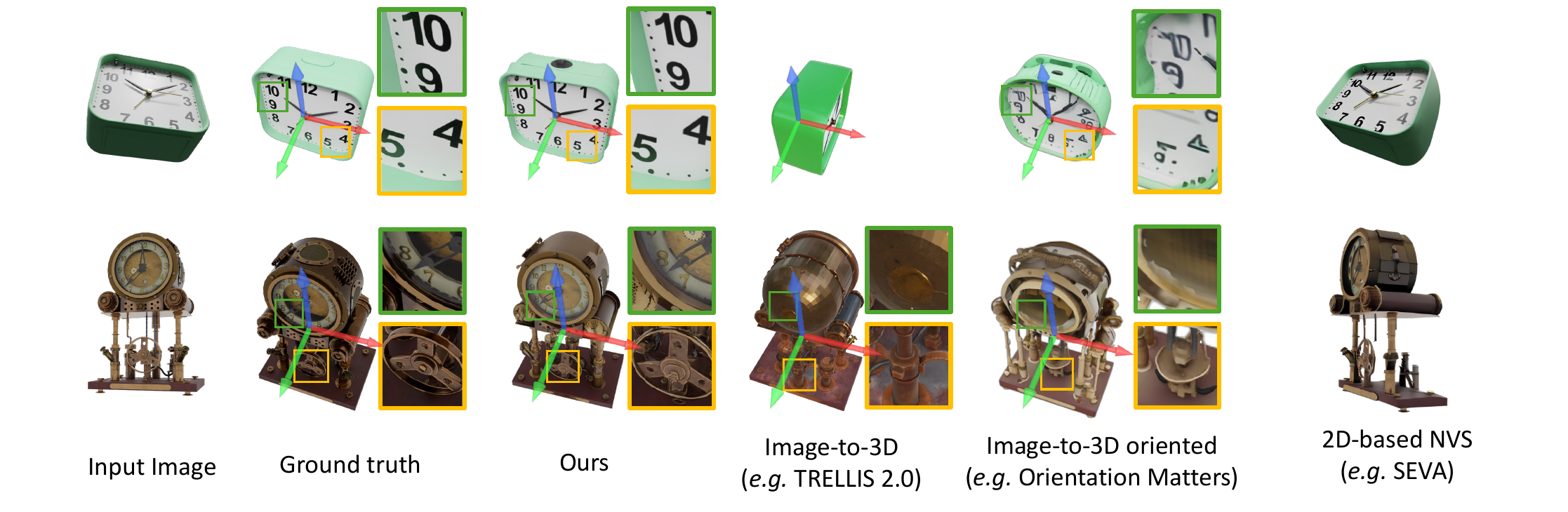}
    \caption{Existing NVS methods fall short in providing global viewpoint control and high-quality appearance: image-to-3D methods (4th column) lack orientation awareness; orientation-aware image-to-3D (5th column) produce coarse textures; and 2D-based NVS (6th column) only provide input-relative camera control. In contrast, our method achieves both global viewpoint control and high image fidelity, with the front face consistently aligned to the canonical frame (visualized by coordinate axes).}
    
    \label{fig:current_limit}
\end{figure}

To mitigate these limitations in NVS viewpoint control, we propose a novel 2D diffusion-based approach for global and intuitive 3D viewpoint control.
We formulate the NVS task in Normalized Object Coordinate Space (NOCS)~\cite{wang2019normalized}, directly conditioning on absolute camera poses in NOCS and thereby eliminating the need for selecting a reference view or constructing a relative world frame from input camera poses. Unlike prior methods that treat NVS as a standalone generation task, we leverage a state-of-the-art instruction-based image editing model~\cite{wu2025qwenimagetechnicalreport} to build an orientation-aware NVS system, inheriting its superior image quality and generalization capabilities.
Specifically, we represent the target NOCS camera pose using spatially-aligned Pl\"ucker raymaps and inject them as camera tokens through an in-context learning strategy with a novel \textit{regional attention mechanism}.
This mechanism ensures that each target camera token attends exclusively to its corresponding target image tokens, preventing cross-view interference.
Since our training data is orientation-aligned, this in-context regional attention enables efficient and consistent viewpoint control across generated views while preserving the advanced 2D image synthesis capabilities of the underlying diffusion model.

\begin{figure}
    \centering
    \includegraphics[width=0.9\linewidth]{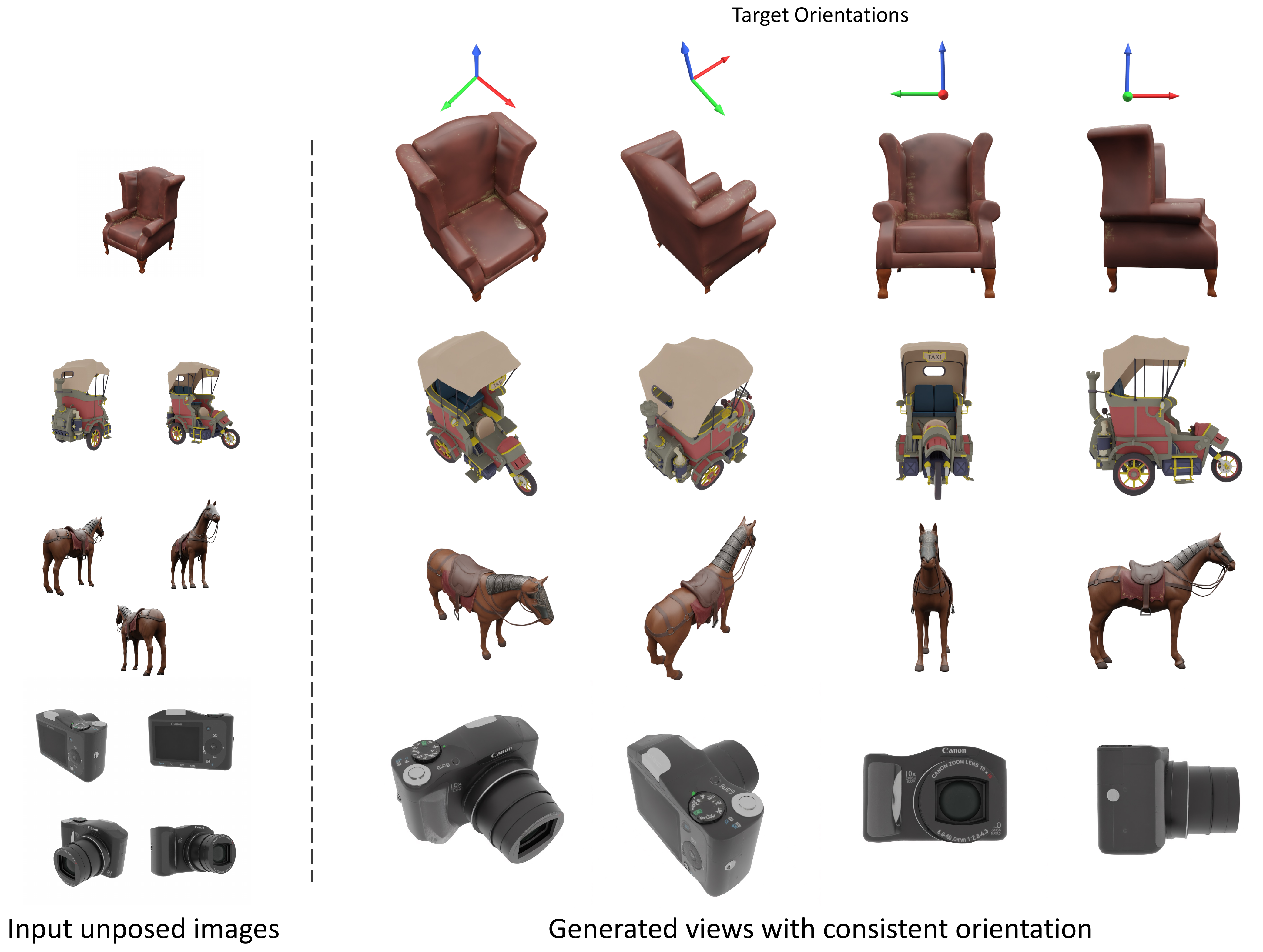}
    \caption{Given input images captured from arbitrary viewpoints (left) and target camera poses (top-right), our model generates high-quality oriented novel views. Notice how the generated images maintain correct global orientation across different object families (\textit{e.g.}, the front of the object aligned with the red axis).
    }
    \label{fig:placeholder}
\end{figure}

Furthermore, we leverage the native text-conditioning capabilities of the underlying editing model to explicitly define the canonical orientation of the input object via natural language (e.g., ``the front of a camera faces the lens side''). This serves a dual purpose: (1) it disambiguates the NOCS definition, as the canonical frame and the target camera pose jointly determine the output viewpoint; and (2) it enables robust generalization to unseen object categories at test time, by allowing users to specify arbitrary canonical orientations through text prompts without retraining.

We train our method on orientation-aligned 3D datasets, introducing a newly curated dataset of 22K objects from Objaverse with no overlap with existing resources, alongside with Objaverse-OA~\cite{lu2025orientation} (14K objects) and ABO~\cite{collins2022abo} (7K objects). Importantly, we augment each object across all datasets with text captions that describe both the object and its front-view definition, enabling language-guided viewpoint control to foster future research.

Our experiments demonstrate that, thanks to our in-context regional attention mechanism and the use of orientation-aligned data, our model achieves state-of-the-art appearance quality for NVS tasks while significantly improving global viewpoint control compared to existing baselines.

\section{Related Work}

Novel view synthesis (NVS) is a foundational computer vision problem. Classical image-based rendering \cite{gortler1996lumigraph, 
levoy1996light} achieved photorealism but required dense multi-view capture. NVS approaches fall into two paradigms: learning 
intermediate 3D representations for rendering, or direct 2D view generation.

\noindent\textbf{Learning 3D Representations.} Generative modeling using GAN\cite{goodfellow2014generative} was extended to learning 3D representations \cite{wu2016learning, Chan2021}. 
Neural Radiance Fields (NeRF)~\cite{mildenhall2020nerf} and subsequent acceleration techniques~\cite{muller2022instant} advanced 3D reconstruction, though per-scene optimization requires dense views. Feed-forward methods like pixelNeRF~\cite{yu2021pixelnerf} introduced generalizable priors for sparse-view settings, while 3D Gaussian Splatting (3DGS)~\cite{kerbl20233d} enabled real-time rendering, spawning efficient sparse-view regressors~\cite{szymanowicz2024splatter, charatan2024pixelsplat, chen2024mvsplat}.
The success of diffusion models~\cite{ho2020denoising, song2019generative} catalyzed a new era of 3D generation, from Score Distillation Sampling~\cite{wang2023score, poole2023dreamfusion, wu2024reconfusion} to multi-view consistent diffusion models~\cite{shi2024mvdream, wang2023imagedream, liu2024syncdreamer, long2024wonder3d, shi2023zero123plus}.
Concurrently, Large Reconstruction Models~\cite{hong2024lrm} directly output 3D representations, and recent 3D native approaches like Trellis and Hunyuan3D~\cite{xiang2024structured,hunyuan3d2025hunyuan3d} encapsulate 3D priors entirely within latent spaces, bypassing intermediate multi-view generation.

\noindent\textbf{Canonical Alignment in 3D Generation:} 
A persistent challenge in 3D generative modeling is the lack of a consistent canonical coordinate space. 
Inspired by single-image 6DOF pose estimation~\cite{wang2019normalized,chen2020category,Peng2018PVNetPV,Sun2022OnePoseOO,Wen2023FoundationPoseU6,wang2025orient}, Orientation Matters~\cite{lu2025orientation} addresses this by leveraging VLMs to align training data and fine-tuning 3D generators to output canonically oriented objects. However, the visual fidelity remains bottlenecked by the expressivity of the underlying 3D generator.

\noindent\textbf{Direct Generative View Synthesis.} 
Instead of explicitly modeling 3D space, several methods synthesize novel views directly via large transformers~\cite{sajjadi2022scene, jin2025lvsm}, image-conditioned diffusion models~\cite{liu2023zero, gao2024cat3d, Szymanowicz2023ViewsetD, tseng2023consistent}, or video diffusion models adapted for camera control~\cite{voleti2024sv3d, wang2024motionctrl, he2024cameractrl, van2024generative, ren2025gen3c, bahmani2025vdd}.
A critical limitation unites these approaches: they parameterize the target camera pose strictly relative to the input image, failing to establish a canonical coordinate system across semantically similar objects. 
Our work addresses this limitation by introducing global viewpoint control through explicit camera conditioning in a normalized object 
coordinate space. Unlike relative pose parameterization, our approach establishes a shared canonical frame across instances, combining 
the flexibility of 2D diffusion models with geometric consistency for global pose control.

\noindent\textbf{Datasets.}
Progress in novel view synthesis and 3D generation has been tightly coupled with 3D object datasets. Google Scanned Objects (GSO) \cite{Downs2022GoogleSO} provides 1,030 high-fidelity scanned household items, 
Toys4K \cite{stojanov21toys4k} offers 4,000 synthetic objects,
while Objaverse~\cite{Deitke2022ObjaverseAU} and Objaverse-XL~\cite{Deitke2023ObjaverseXLAU} scaled 3D data to 800K+ and 10M+ objects, respectively.
To address inconsistent orientations in Objaverse, \cite{lu2025orientation} proposed Objaverse-OA, filtering 14K orientation-aligned samples from Objaverse-LVIS. 
However, it still contains objects with imprecise orientations, ambiguous canonical views, or overly simplified textures.
To complement this effort, we curate an additional 22K high-quality objects from Objaverse with no overlap with Objaverse-OA, each with a clear front face, consistent orientation, and detailed textures.

\section{Global Pose Control}
\label{sec:globa-pose-control}
\subsection{Problem Definition}
Given $M$ arbitrary input view images $\mathbf{I}^{\text{src}} \in \mathbb{R}^{M\times H \times W \times 3}$, $N$ target camera poses $\mathbf{\Pi}^{\text{tgt}} \in \text{SE}(3)^{N}$ defined in Normalized Object Coordinate Space (NOCS), and a text description $\mathbf{y}$ that specifies how the object is oriented within NOCS, our goal is to learn the conditional distribution of $N$ target views $\mathbf{I}^{\text{tgt}}$:
\begin{equation}
    \mathbf{I}^{\text{tgt}} \sim p_\theta(\mathbf{I}^{\text{tgt}} \mid \mathbf{I}^{\text{src}}, \mathbf{\Pi}^{\text{tgt}}, \mathbf{y}),
\end{equation}
where $p_\theta$ is parameterized by a diffusion model.

We formulate this as an instruction-based image editing problem, as NVS can be naturally viewed as a spatial transformation of the input image. Accordingly, we build our method upon a state-of-the-art image editing model~\cite{wu2025qwenimagetechnicalreport} to inherit its strong image manipulation capabilities. However, as we demonstrate in our ablation study (Section~\ref{sec:ablation}), relying solely on text instructions is insufficient for robust \textit{global} viewpoint control, due to the inherent complexity and ambiguity of specifying precise 3D camera poses through language alone.

\begin{figure}[t]
    \centering
    \includegraphics[width=1.0\linewidth]{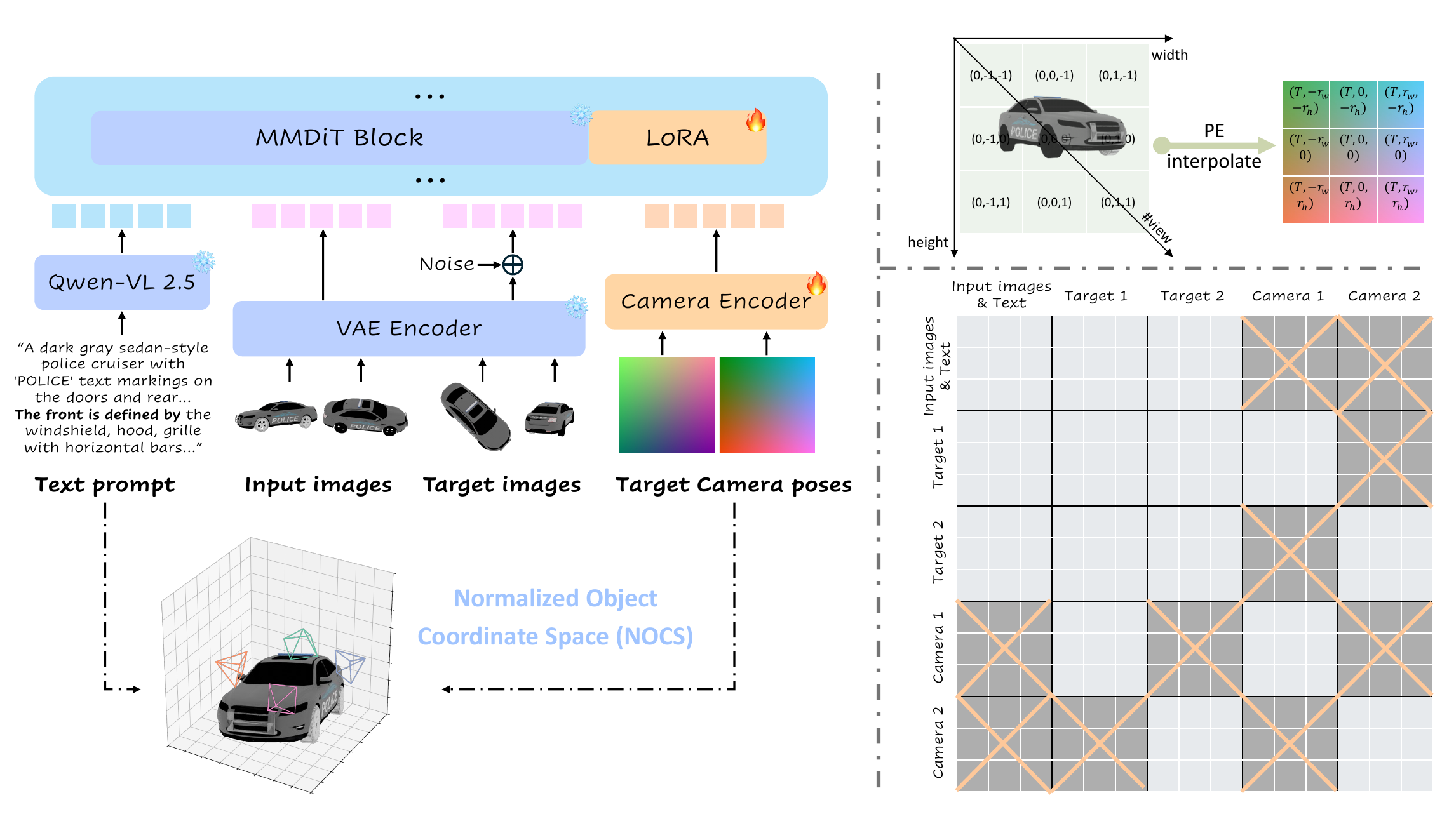}
    \caption{\textbf{Overview.} 
    \textit{Left:} Target camera poses are encoded as Pl\"ucker ray map tokens and fed into a LoRA-adapted MMDiT alongside image tokens. A text prompt defines the object's canonical orientation in NOCS.
    \textit{Top-right:} Target images and ray map tokens are packed along the frame axis. The position embeddings of ray map tokens are interpolated to spatially align with the corresponding image tokens.
    \textit{Bottom-right:} Regional attention ensures each camera token attends only to its corresponding target view, while all image tokens retain full mutual attention.
    }
    \label{fig:pipeline}
\end{figure}

Figure \ref{fig:pipeline} and the remainder of this section describe our architecture, which introduces: in-context camera conditioning tailored for global camera control (Section \ref{sec:in-context-camera}), an efficient and robust training scheme (Section \ref{sec:efficient-training}), and a new aligned dataset (Section \ref{sec:dataset}). 

\subsection{In-context Camera Conditioning}
\label{sec:in-context-camera}
We introduce camera conditioning through an in-context learning method \cite{lhhuang2024iclora}, which has been proven effective for DiT-based conditional generation.
We describe below our camera representation, how we extend the architecture to multi-view generation, and how we ensure correct camera--view association.

\noindent\textbf{Camera Representation.}
We represent each target camera pose as a Pl\"ucker ray map, encoding camera rays as pixel-aligned 6D vectors that jointly capture 2D spatial correspondence and 3D ray geometry. Compared to attention-level relative encodings~\cite{kong2024eschernet,miyato2023gta,li2025cameras}, this absolute, pixel-aligned representation naturally suits our NOCS-based formulation, where target cameras are specified in a global coordinate frame.
Each ray map is projected into token space via a linear layer, producing camera tokens that are appended to the image tokens along the frame dimension and processed jointly by the MMDiT.
We fine-tune only lightweight LoRA~\cite{hu2022lora} adapters on the MMDiT layers to efficiently adapt the pretrained model while preserving its image manipulation capabilities.

\noindent\textbf{Frame-axis Multi-view Packing.}
To generate multiple target views simultaneously, existing methods spatially concatenate outputs into a single wide image~\cite{liang2025UnitTEX, feng2025seed3d10imageshighfidelity}, producing target tokens of shape $H \times NW$ against reference tokens of shape $H \times W$. This resolution mismatch disrupts the pixel-level spatial correspondence that attention mechanisms rely on.
Instead, we leverage the 3D RoPE employed in Qwen-Image-Edit, which encodes positional information along three dimensions: frame, height, and width. In the original editing framework, the frame dimension serves to distinguish between the images before and after editing. We naturally extend this design by packing all target views, together with their corresponding ray maps, as separate entries along the \emph{frame} axis while preserving the original $H \times W$ aspect ratio for each view. This naturally reuses the base model's frame dimension and maintains consistent spatial alignment between reference and target tokens.

\noindent\textbf{Compact Ray Maps with RoPE Interpolation.}
Although Pl\"ucker ray maps are pixel-aligned by construction, camera pose is inherently a low-dimensional signal that does not require pixel-level granularity. Encoding ray maps at full image resolution (\eg, $1024 \times 1024$) would introduce unnecessarily long token sequences without commensurate benefit. We therefore fix the ray map resolution to $256 \times 256$ regardless of the target image resolution.
To maintain spatial alignment with higher-resolution image tokens under the 3D RoPE, we adopt a position-aware interpolation strategy inspired by~\cite{zhang2025easycontroladdingefficientflexible}. As illustrated in the top-right of Figure~\ref{fig:pipeline}, a ray map token at grid position $(i, j)$ is assigned RoPE coordinates $(i \cdot r_h,\; j \cdot r_w)$, where $r_h = H / 256$ and $r_w = W / 256$.
This ensures that spatially co-located ray map and image tokens receive matching positional encodings, preserving their correspondence at reduced computational cost.

\noindent\textbf{Regional Attention.}
With frame-axis packing, all entries, including multiple target views, reference images, and their corresponding ray maps, share the same $H \times W$ spatial grid and differ only along the frame dimension. This introduces an unintended side effect in the 3D RoPE attention: tokens from different entries that occupy the same spatial position have identical height and width indices, making their relative distance determined solely by the frame offset. For example, with two target views and two camera ray maps packed as $[\mathbf{z}^{\text{tgt}}_1, \mathbf{z}^{\text{tgt}}_2, \mathbf{z}^{\text{cam}}_1, \mathbf{z}^{\text{cam}}_2]$, a patch in $\mathbf{z}^{\text{cam}}_1$ may be closer in frame index to $\mathbf{z}^{\text{tgt}}_2$ than to $\mathbf{z}^{\text{tgt}}_1$, causing stronger attention between $\mathbf{z}^{\text{cam}}_1$ and the wrong target view and violating the intended camera-view correspondence.
We resolve this with a regional attention mask, as shown in Figure~\ref{fig:pipeline} (bottom-right), that enforces correct camera-view binding: each camera token attends \emph{only} to its corresponding target view, and conversely, each target view is restricted to attend \emph{only} to its matched camera tokens among all camera tokens.
All image tokens, both reference and target, retain full mutual attention, enabling multi-view information exchange while preventing cross-view camera leakage. This design also saves computation by masking out unrelated camera-view attention pairs.

\subsection{Efficient and Robust Training}
\label{sec:efficient-training}
\noindent\textbf{View Sampling.}
Each training sample consists of $T = M + N$ views in total, where $M$ is the number of input views and $N$ is the number of target views. Rather than fixing these counts, we dynamically sample them per iteration to expose the model to diverse task configurations and difficulty levels.

We first sample the total number of views $T$ from a power-weighted distribution over $\{T_{\min}, \dots, T_{\max}\}$:
\begin{equation}
    p(T{=}t) \;\propto\; (t - T_{\min} + \epsilon)^{k},
\end{equation}
where $k{=}1.5$ and $\epsilon{=}1$. This biases sampling toward larger $T$, presenting the model with more challenging multi-view configurations.

Given $T$, we then sample the number of input views $M$ from a truncated exponential distribution over $\{1, \dots, T{-}1\}$, following~\cite{liu2026kaleido}:
\begin{equation}
    p(M{=}m) \;\propto\; e^{-\lambda m}, \quad \lambda = \ln 2,
\end{equation}
which halves the probability as $m$ increases. Since more reference views make the task increasingly constrained and easier, this biases toward fewer input views, encouraging the model to learn stronger priors from limited observations.

To ensure balanced workload across GPUs, we bucket training samples by their $(M, N)$ configuration and enforce a uniform $(M, N)$ assignment within each iteration, so that all devices process sequences of equal length.

\noindent\textbf{Canonical Front View Anchoring.}
Our text prompt explicitly defines the front-facing orientation of each object in NOCS. To ground this description into a concrete visual reference, we include the canonical front view among the generated targets. This front view is generated jointly with all other target views in a single diffusion process, rather than as a separate preliminary step. Nevertheless, it encourages the model to first resolve the object's canonical orientation before reasoning about the remaining viewpoints, acting as an implicit intermediate step akin to chain-of-thought prompting~\cite{wei2022chain,wiedemer2025video}. As shown in our ablation (Section~\ref{sec:ablation}), this simple design notably improves pose accuracy for geometrically challenging objects (\eg, narrow or thin shapes such as swords).

\noindent\textbf{Loss Function.}
We adopt a flow matching objective for training. Given latent $x_0 = \mathcal{E}(\mathbf{I}^{\text{tgt}})$ encoded by the VAE and noise $x_1 \sim \mathcal{N}(0, \mathbf{I})$, the intermediate latent at timestep $t$ is $x_t = t x_0 + (1-t) x_1$, with velocity $v_t = x_0 - x_1$. The model is trained to predict this velocity via:
\begin{equation}
    \mathcal{L} = \mathbb{E}_{(x_0, h) \sim \mathcal{D},\, x_1,\, t} \left\| v_\theta(x_t, t, h) - v_t \right\|^2,
\end{equation}
where $h$ denotes conditioning from input images, text, and camera poses.

\subsection{Dataset}
\label{sec:dataset}
\vspace{-10pt}
\begin{wrapfigure}[14]{l}{0.40\textwidth}
\vspace{-15pt}
  \centering
  \includegraphics[width=0.4\textwidth]{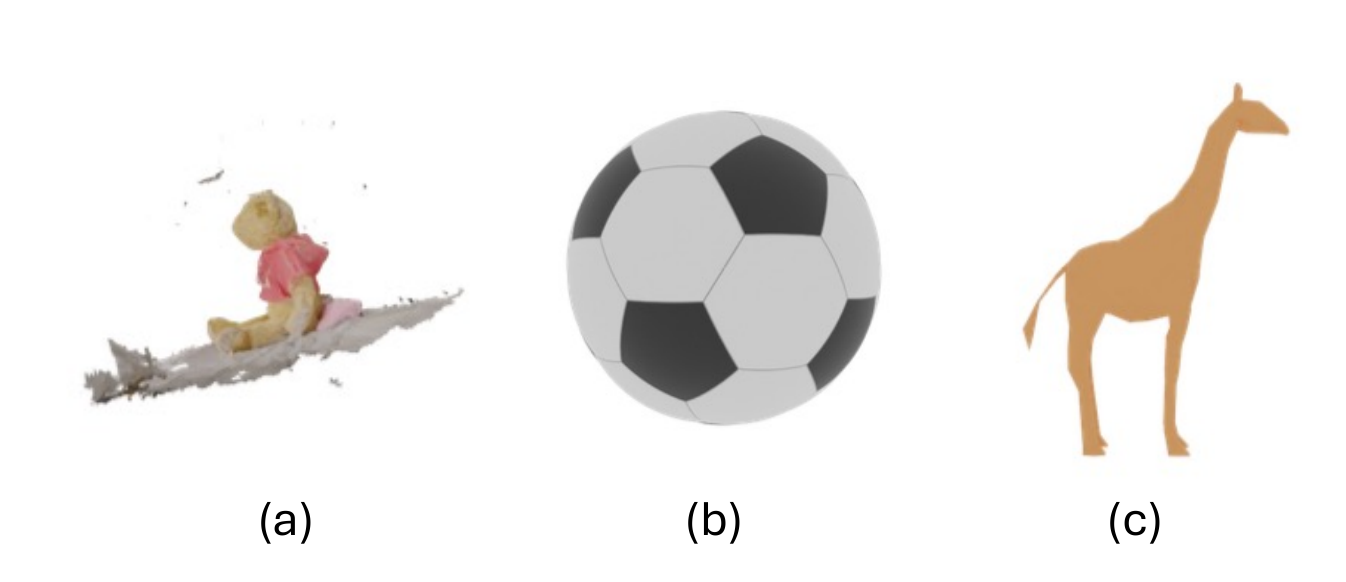}
    \caption{\textbf{Common failure modes in Objaverse-OA.} \textbf{(a)}~objects with imprecise orientations (\eg, tilted or rotated off-axis), \textbf{(b)}~objects with no recognizable or ambiguous front views, and \textbf{(c)}~objects with over-simplified textures.}
  \label{fig:dataset_limit}
\end{wrapfigure}
Our approach requires canonically aligned, high-quality 3D objects with front-face annotations at training time. Unfortunately, no existing dataset fully satisfies these requirements: Objaverse-OA~\cite{lu2025orientation} filters 14K aligned objects from Objaverse-LVIS, but we observe that many samples are tilted (yaw or pitch offset), lack a valid front face, or exhibit over-simplified textures (Figure~\ref{fig:dataset_limit}).
ABO~\cite{collins2022abo} provides 8K models in a canonical coordinate space, but its categories are limited to household objects.
We therefore propose a robust curation pipeline to assemble a larger and cleaner collection from Objaverse~\cite{Deitke2022ObjaverseAU}, enriched with textual front-face definitions.
To ensure high quality and canonical alignment, we combine Orient-Anything-V2~\cite{wangorientv2} with VLM-based~\cite{claude45} verification,
prioritizing strict filtering to minimize erroneous samples while maintaining category diversity.

We start from the Objaverse subset of HY3D-Bench~\cite{hunyuan3d2026hy3dbenchgeneration3dassets}, which provides 70K objects filtered for high texture quality from Objaverse~\cite{Deitke2022ObjaverseAU}. However, HY3D-Bench does not enforce canonical orientation or guarantee a valid front face. We therefore apply a multi-stage curation pipeline on top of this pool.

We first remove tilted and skewed objects using Orient-Anything-V2~\cite{wangorientv2}.
Specifically, we retain only objects where the predicted elevation degree is less than 8° and the azimuth degree falls within a 5° tolerance of the canonical angles (0°, 90°, 180°, or 270°).
This ensures that objects are properly upright and aligned to the principal axes.
To further refine our selection, we employ a VLM to verify whether objects are properly placed or tilted, providing an additional layer of quality control beyond geometric constraints alone.

We then filter objects based on the number of valid of \textit{frontal} faces, following the definition and implementation of Orient-Anything-V2 \cite{wangorientv2}.
Objects with 0 valid faces (indicating no recognizable canonical views, for example a soccer ball) or 4 valid faces (suggesting ambiguous or symmetric geometry, for example a rectangle basket) are excluded.
The resulting dataset consists of objects with 1 or 2 valid faces, which provide clear canonical orientations suitable for training.

For front face annotation, Orient-Anything-V2 successfully labels approximately 21K objects by selecting the view among the 
four horizontal candidates with azimuth closest to 0° and within a 5° tolerance.
For the remaining ~1K challenging cases, such as elongated 
objects like swords that may have two equally valid front faces, we employ VLM-based annotation to resolve ambiguities. 

Finally, we annotate every sample with an overall object description and a textual front-face definition using the same VLM. The front-face definition specifies which side of the object constitutes the front, thereby anchoring the NOCS frame (\eg, ``the front is defined by the main gun barrel pointing forward''; ``the front is defined by the side with the handlebars and grips facing the viewer'').
Through this pipeline we curate \textbf{22K} canonically oriented objects with front-face labels from the initial 70K pool, complementing existing datasets such as Objaverse-OA and ABO while significantly broadening category coverage.

\section{Experiments}

\subsection{Implementation Details} 
We build upon Qwen-Image-Edit-2509~\cite{wu2025qwenimagetechnicalreport} as our base model and apply LoRA~\cite{hu2022lora} with rank 64 to efficiently adapt the pretrained weights. We use the AdamW optimizer with a learning rate of $1 \times 10^{-4}$ and a per-GPU batch size of 1 on 8 NVIDIA H100 GPUs. We adopt a coarse-to-fine training strategy: the model is first trained at $512{\times}512$ resolution for 20k steps to learn camera control and multi-view consistency, then fine-tuned at $1024{\times}1024$ for 15k steps to recover high-frequency details. Our training data comprises Objaverse-OA~\cite{lu2025orientation}, Amazon Berkeley Objects (ABO)~\cite{collins2022abo}, and our custom aligned dataset (Section~\ref{sec:dataset}).

We evaluate on two publicly available benchmarks: Google Scanned Objects (GSO)~\cite{Downs2022GoogleSO} and Toys4k~\cite{stojanov21toys4k}, selecting 100 objects with clearly identifiable front faces from each. Metrics include PSNR, SSIM, LPIPS~\cite{zhang2018unreasonable} and CLIP~\cite{radford2021learning} scores.

We compare against two categories of methods: (1)~\emph{image-to-3D methods}: Orientation Matters~\cite{lu2025orientation}, Cupid~\cite{huang2025cupid}, Hunyuan3D 2.1~\cite{hunyuan3d2025hunyuan3d}, and Trellis-2~\cite{xiang2025trellis2}, which generate 3D assets from a single input image; and (2)~\emph{Generative NVS methods}: EscherNet~\cite{kong2024eschernet} and SEVA~\cite{zhou2025stable}, which we evaluate with 1, 2, 3, and 5 input views to assess performance across varying input configurations.

\subsection{Comparisons with 3D methods}
\begin{table*}[tb]
\captionof{table}{\textbf{Quantitative Comparison with image-to-3D methods.}}
\label{tab:quality_3d}
\centering
\resizebox{.95\textwidth}{!} 
{
\begin{tabular}{l|cccc|cccc}
\toprule
& \multicolumn{4}{c|}{GSO~\cite{Downs2022GoogleSO}} & \multicolumn{4}{c}{Toys4k~\cite{stojanov21toys4k}} \\
& PSNR$\uparrow$ & SSIM$\uparrow$ & LPIPS$\downarrow$ & CLIP$\uparrow$ & PSNR$\uparrow$ & SSIM$\uparrow$ & LPIPS$\downarrow$ & CLIP$\uparrow$ \\
\midrule
Cupid~\cite{huang2025cupid}& 15.32 & 0.837 & 0.186 & 91.3 & 16.67 & 0.878 & 0.132 & 94.0 \\
Orientation Matters \cite{lu2025orientation} & 16.13 & 0.844 & 0.178 & 89.8 & 16.94 & 0.879 & 0.135 & 91.6 \\
TRELLIS2\cite{xiang2025trellis2} & 15.77 & 0.837 & 0.187 & 90.5 & 15.64 & 0.862 & 0.156 & 91.2 \\
HY3D-2.1\cite{hunyuan3d2026hy3dbenchgeneration3dassets} & 15.21 & 0.837 & 0.196 & 89.3 & 16.67 & 0.873 & 0.145 & 91.4 \\
\midrule
Ours & \textbf{17.25} & \textbf{0.855} & \textbf{0.163} & \textbf{92.3} & \textbf{18.24} & \textbf{0.892} & \textbf{0.116} & \textbf{94.2} \\
\bottomrule
\end{tabular}
}
\end{table*}

\begin{figure}[tb]
    \centering
    \includegraphics[width=0.98\linewidth]{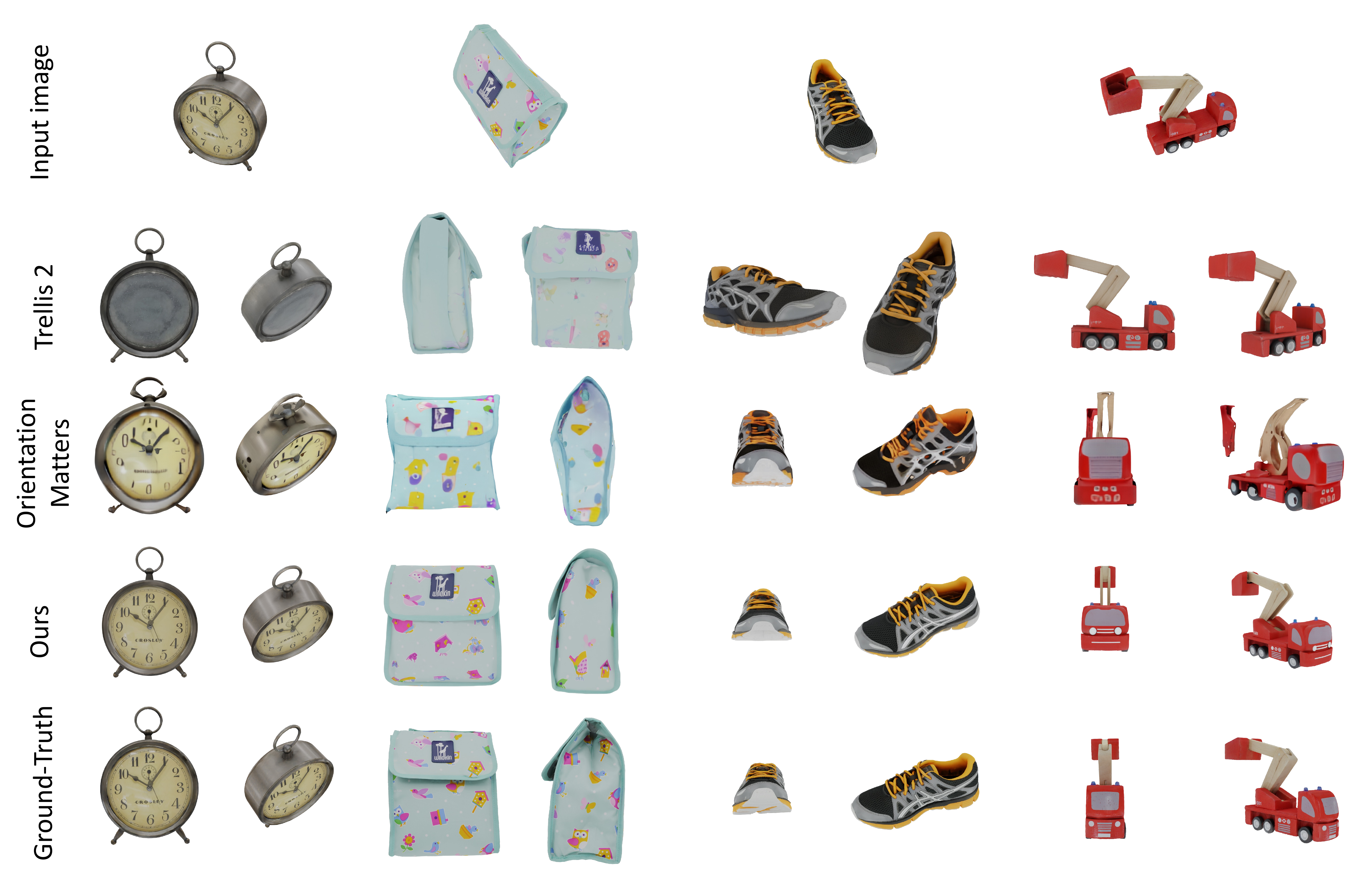}
    \caption{\textbf{Qualitative comparison with image-to-3D methods.} For each object, we show two target views: the canonical front view (left) and an arbitrary viewpoint (right). Our approach faithfully preserves fine-grained textures (e.g., clock face numerals) and structural details (e.g., shoe laces), whereas 3D methods suffer from texture loss, blurriness, or geometric distortions. Best viewed zoomed in.}
    \label{fig:compare_3d}
\end{figure}

Table~\ref{tab:quality_3d} presents a quantitative comparison between our method and state-of-the-art 3D generation approaches, all using a single-view image as input. Our method consistently outperforms all baselines across both benchmarks and all metrics.
The improvements are particularly pronounced on Toys4k, which contains challenging categories such as vehicles and shoes with intricate geometry and texture patterns.

We attribute the limitations of existing 3D methods to two primary factors: \emph{orientation inconsistency} and \emph{texture degradation}. Trellis-2 and Hunyuan3D 2.1 frequently produce 3D assets with incorrect canonical orientations, leading to misaligned novel views. While Orientation Matters mitigates this by fine-tuning on orientation-aligned 3D data, and Cupid jointly predicts canonical geometry and camera pose, both still suffer from significant texture quality loss inherent to the 3D generation pipeline.
By operating directly in image space, our method avoids the information bottleneck of 3D methods and preserves significantly richer visual detail, as shown in Figure~\ref{fig:compare_3d}.

\subsection{Comparisons with NVS methods}
\begin{figure}[tb]
    \centering
    \includegraphics[width=0.98\linewidth]{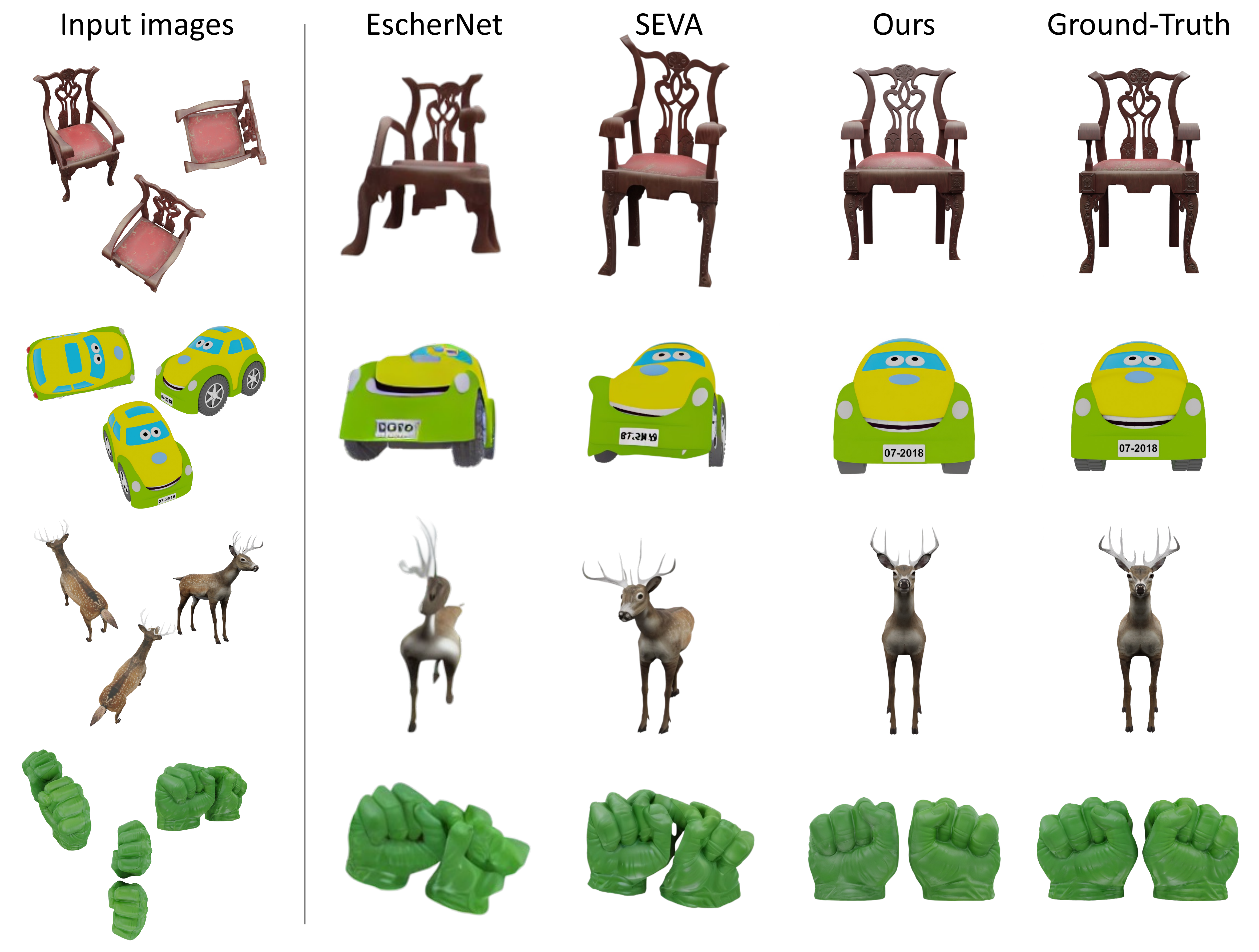}
    \caption{\textbf{Qualitative comparison with generative NVS methods.} We show the generated front view for each method. Both EscherNet and SEVA produce results with noticeably incorrect orientations due to errors in the estimated input camera poses. In contrast, our method operates directly in canonical space without requiring input pose estimation, generating accurately oriented front views.}
    \label{fig:compare_nvs}
\end{figure}

\begin{table*}[tb]
\caption{\textbf{Quantitative Comparison with generative NVS methods}.}
\label{tab:compare_nvs}
\centering
\small
\setlength{\tabcolsep}{3pt}

\begin{tabular}{l l *{4}{c} | *{4}{c}}
\toprule
\multirow{2}{*}{} & 
& \multicolumn{4}{c}{GSO}
& \multicolumn{4}{c}{Toys4k} \\
\cmidrule(lr){3-6} \cmidrule(lr){7-10}
& \#Input Views 
& 1 & 2 & 3 & 5 
& 1 & 2 & 3 & 5 \\
\midrule

EscherNet~\cite{kong2024eschernet}
& PSNR$\uparrow$  & 14.90 & 17.25 & 17.58 & 17.97 & 15.32 & 15.36 & 15.77 & 16.26 \\
& SSIM$\uparrow$  & 0.832 & 0.855 & 0.860 & 0.865 & 0.861 & 0.864 & 0.871 & 0.880 \\
& LPIPS$\downarrow$ & 0.230 & 0.187 & 0.180 & 0.172 & 0.196 & 0.192 & 0.181 & 0.169 \\
\midrule

SEVA~\cite{zhou2025stable}
& PSNR$\uparrow$  & 15.20 & 15.85 & 16.13 & 16.39 & 15.87 & 15.85 & 16.27 & 16.66 \\
& SSIM$\uparrow$  & 0.834 & 0.839 & 0.844 & 0.849 & 0.868 & 0.868 & 0.874 & 0.880 \\
& LPIPS$\downarrow$ & 0.188 & 0.181 & 0.176 & 0.170 & 0.147 & 0.152 & 0.143 & 0.134 \\
\midrule

Ours
& PSNR$\uparrow$  & 17.25 & 18.11 & 18.52 & 19.07 & 18.24  & 18.77 & 19.17 & 19.68 \\
& SSIM$\uparrow$  & 0.855 & 0.863 & 0.866 & 0.871 & 0.892 & 0.895 & 0.897 & 0.902 \\
& LPIPS$\downarrow$ & 0.163 & 0.142 & 0.136 & 0.128 & 0.116 & 0.107 & 0.100 & 0.093 \\
\bottomrule
\end{tabular}
\end{table*}

We compare our method with state-of-the-art generative NVS methods, EscherNet~\cite{kong2024eschernet} and SEVA~\cite{zhou2025stable}, under varying numbers of input views ($M{=}1, 2, 3, 5$). 
Both methods operate in a relative coordinate frame and require known camera poses for all input images. Since our evaluation setting assumes unposed inputs, we use Orientation-Anything-V2~\cite{wangorientv2} to estimate input orientations in canonical space and convert them to the required camera poses.

Quantitative results in Table~\ref{tab:compare_nvs} show that our method consistently outperforms both baselines across all input configurations. A key reason is the \emph{error accumulation} inherent to relative-pose-based methods: inaccuracies in the estimated input poses propagate directly into the generated target views, causing significant quality degradation. This issue is exacerbated by the entangled nature of Euler angles, where Orientation-Anything-V2 particularly struggles with roll estimation, compounding errors across views. By operating directly in the NOCS space, our method eliminates the dependency on input camera pose estimation entirely, making it inherently robust to this source of error.

\subsection{Ablation Study}
\label{sec:ablation}
\begin{figure}[tb]
    \centering
    \includegraphics[width=0.98\linewidth]{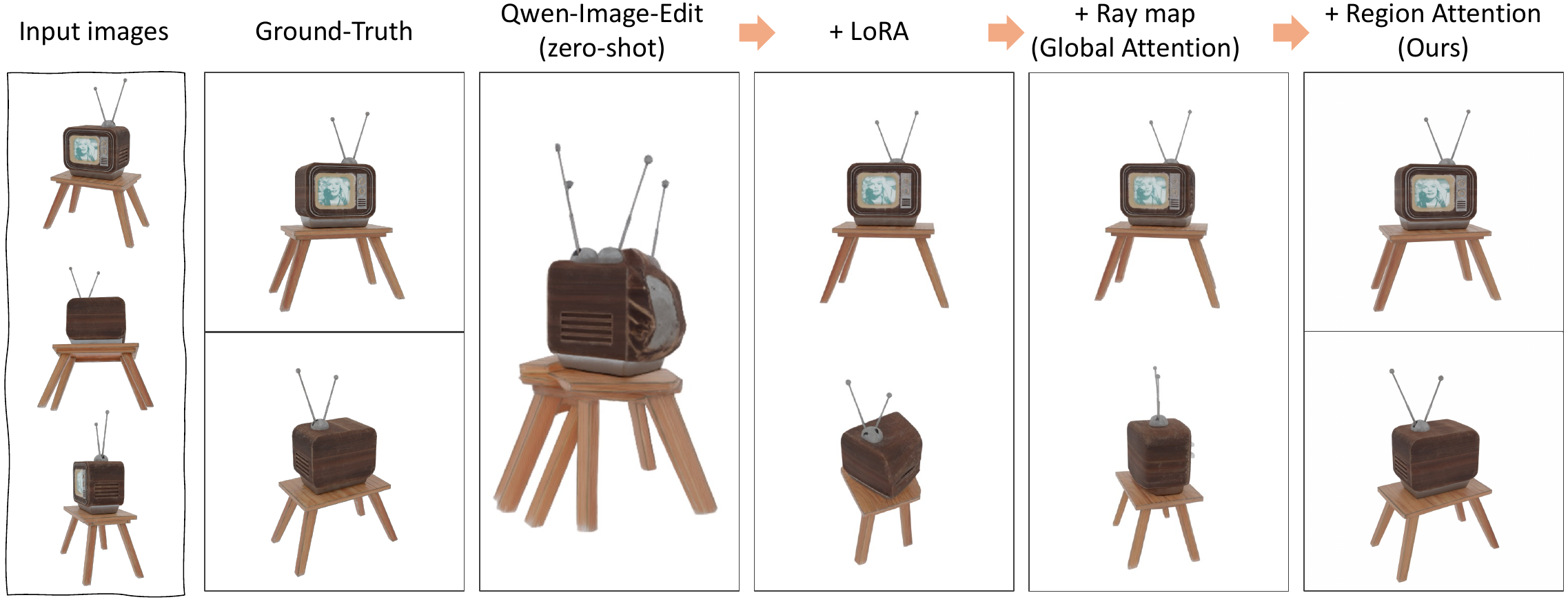}
    \caption{\textbf{Ablation study on camera conditioning.} Qwen-Image-Edit cannot robustly produce multi-view layouts without fine-tuning. LoRA enables layout generation, but text-only camera descriptions lack precision. Plücker raymap tokens provide explicit geometric conditioning, yet full attention causes cross-view leakage. Regional attention restricts each raymap to its corresponding view, eliminating leakage and achieving accurate pose control.}
    \label{fig:ab_cam}
\end{figure}

In this section we ablate different design choices and components of our method.
\noindent\textbf{Camera Conditioning.}
Table \ref{tab:ablation-camera} and Figure~\ref{fig:ab_cam} ablate camera conditioning design choices.
As baseline, we use off-the-shelf Qwen-Image-Edit with structured prompts specifying NOCS definitions and camera parameters (see Supplementary). Despite explicit specification, the model fails to reliably generate multi-view outputs with correct view counts.
LoRA training with structured prompts enables multi-view generation but produces substantial pose deviations. We attribute this to the mismatch between discrete text tokens and continuous SE(3) camera parameters—geometric structure is difficult to encode through text alone.
Replacing text with Plücker raymaps provides geometric representation. An encoder maps raymaps to camera tokens concatenated with image tokens. However, naïve fusion fails: appending camera tokens causes cross-view entanglement and ineffective injection (Section \ref{sec:in-context-camera}).
Regional attention explicitly binds each camera token to its spatial region, successfully injecting camera information while preventing cross-view interference.
Training and evaluation use a subset of our dataset

\begin{table}[tb]
  \caption{\textbf{Ablation study on camera conditioning.}}
  \label{tab:ablation-camera}
  \centering
  \begin{tabular}{l|ccc}
    \toprule
    Method & PSNR $\uparrow$ & SSIM $\uparrow$ & LPIPS $\downarrow$ \\
    \midrule
    Baseline & 12.48 & 0.750 & 0.413 \\
    Baseline + LoRA & 16.78 & 0.870 & 0.184 \\
    Baseline + LoRA + Camera tokens w/ Global Attention & 16.87 & 0.870 & 0.180  \\
    Baseline + LoRA + Camera tokens w/ Region Attention & \textbf{20.80} & \textbf{0.899} & \textbf{0.111}  \\
  \bottomrule
  \end{tabular}
\end{table}

\begin{figure}[tb]
    \centering
    \includegraphics[width=0.98\linewidth]{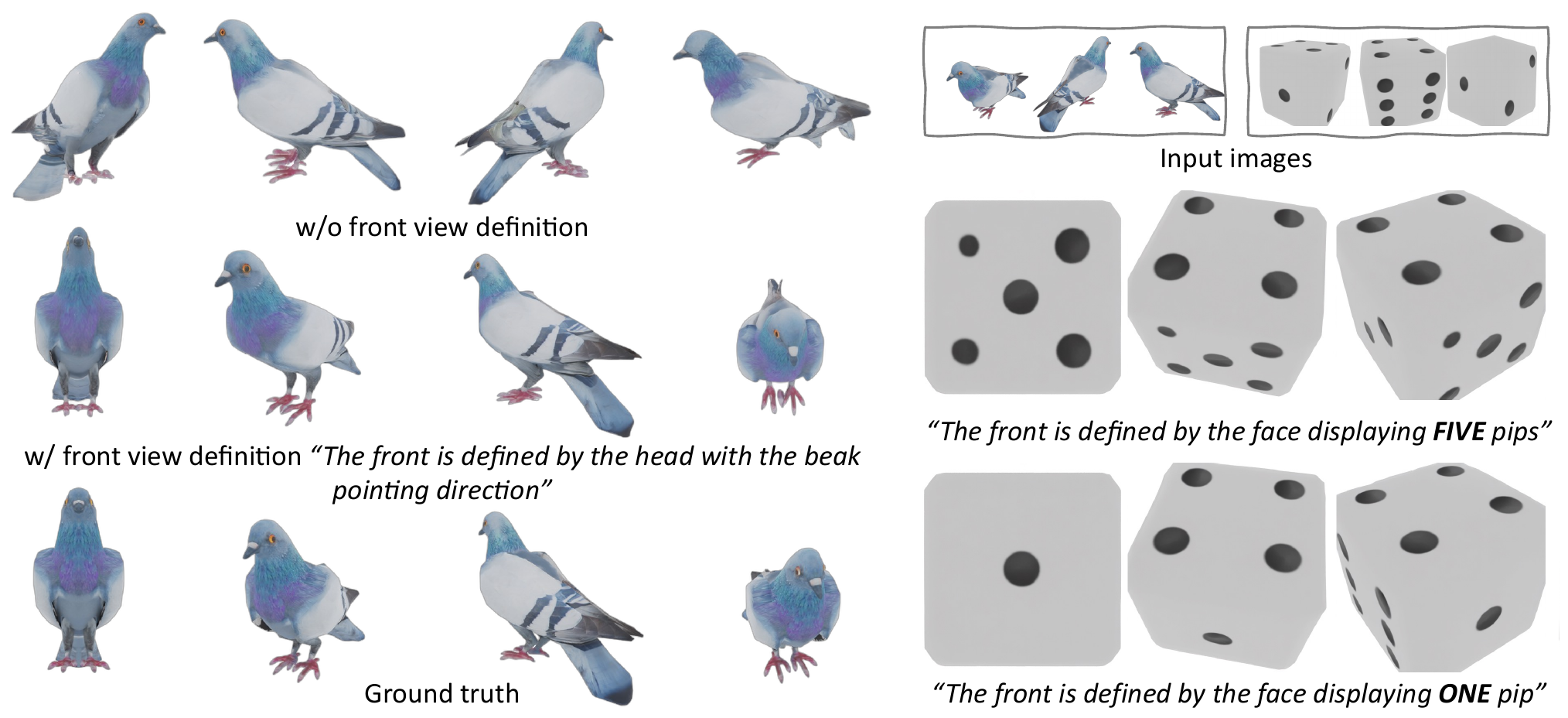}
    \caption{\textbf{Ablation study on front view definition.} With front face definition, our model robustly generalizes to unseen categories (left) and enables test-time redefinition of front view (right).}
    \label{fig:ab_text}
\end{figure}

\noindent\textbf{Front Face Definition.} 
To ablate the impact of front face definition, we train our model only on the ABO \cite{collins2022abo} dataset, which contains household 
objects, with and without front view definition. 
As shown Figure~\ref{fig:ab_text} (left), front view definition helps the model generalize to unseen categories such as a pigeon.
Importantly, the advantages of this front face definition strategy go beyond generalization capabilities: we can also define arbitrary front 
views and normalized spaces as desired, and our model reacts accordingly.
In Figure~\ref{fig:ab_text} (right), we show how redefining the 
front face of a dice results in correctly-aligned NVS control that follows the input text definition.
Such test-time robust control of the global canonical frame is something existing models cannot provide.  

\begin{figure}[tb]
    \centering
    \includegraphics[width=0.98\linewidth]{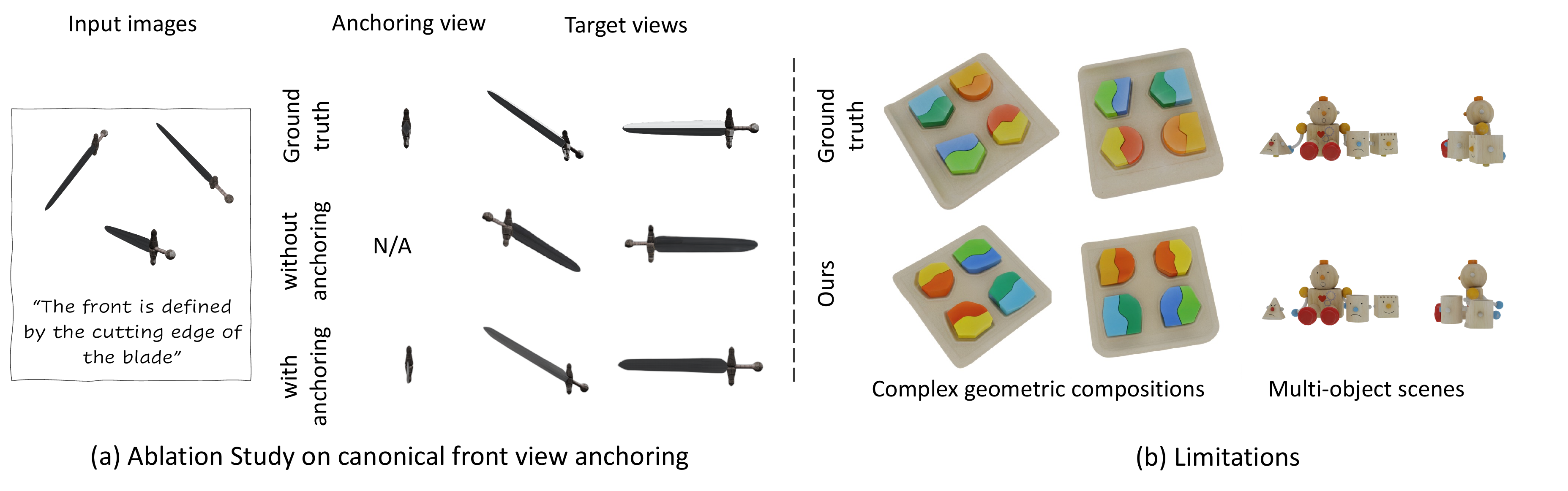}
    \caption{\textbf{Ablation study on canonical front view anchoring and Limitations.}}
    \label{fig:limitations}
\end{figure}

\noindent\textbf{Canonical Front View Anchoring}
We ablate the effect of including the canonical front view among the generated targets. As shown in Figure~\ref{fig:limitations}(a), this design is particularly beneficial for geometrically challenging objects such as swords, where the narrow, elongated shape makes it difficult to resolve the canonical orientation from arbitrary viewpoints alone. Without front anchoring, the model confuses the front–back orientation, producing views with incorrect pose (\eg, the blade appears flattened or reversed). By generating both the front view and the target view, the model establishes a consistent spatial reference that guides the remaining target views toward correct orientations.

\FloatBarrier

\section{Conclusion}

We presented a novel approach for novel view synthesis with precise global viewpoint control in Normalized Object Coordinate Space (NOCS). By formulating NVS as an image editing problem, our method inherits the superior image quality of state-of-the-art editing models, while our in-context camera conditioning with regional attention ensures correct camera-view binding. Text-based canonical orientation descriptions disambiguate the NOCS definition and enable generalization to unseen categories without retraining. To support training, we curated a high-quality dataset of 22K canonically aligned objects with front-face text definitions, complementing existing resources. Extensive experiments demonstrate that our method achieves high image quality and accurate global viewpoint control from arbitrary unposed images, significantly outperforming both image-to-3D and generative NVS baselines.

\noindent\textbf{Limitations and Future Work.} Despite its strong performance, our method has several limitations. First, our current approach assumes fixed camera intrinsics across all views, limiting its flexibility for applications requiring varying focal lengths. Second, as shown in Figure~\ref{fig:limitations}~(left), although text descriptions help disambiguate the canonical NOCS frame, objects with complex geometric compositions remain challenging. Third, our method is primarily designed for single-object scenarios. When the input contains multiple objects with distinct orientations, performance degrades. Finally, integrating our globally controlled novel views as a drop-in module for downstream tasks such as robotic manipulation and content creation presents a promising direction for future work.

\bibliographystyle{splncs04}
\bibliography{main}

@String(CVPR  = {IEEE/CVF Conference on Computer Vision and Pattern Recognition (CVPR)})

@String(ICCV  = {International Conference on Computer Vision (ICCV)})

@String(ECCV  = {European Conference on Computer Vision (ECCV)})

@String(NeurIPS = {Conference on Neural Information Processing Systems (NeurIPS)})

@String(ICML  = {International Conference on Machine Learning (ICML)})

@String(ICLR  = {International Conference on Learning Representations (ICLR)})

@String(TOG   = {ACM Trans. Graph.})

@inproceedings{
shi2024mvdream,
title={{MVDream: Multi-view Diffusion for 3D Generation}},
author={Yichun Shi and Peng Wang and Jianglong Ye and Long Mai and Kejie Li and Xiao Yang},
booktitle=ICLR,
year={2024},
url={https://openreview.net/forum?id=FUgrjq2pbB}
}

@inproceedings{Deitke2022ObjaverseAU,
  title={{Objaverse: A Universe of Annotated 3D Objects}},
  author={Matt Deitke and Dustin Schwenk and Jordi Salvador and Luca Weihs and Oscar Michel and Eli VanderBilt and Ludwig Schmidt and Kiana Ehsani and Aniruddha Kembhavi and Ali Farhadi},
  booktitle = CVPR,
  year={2022},
  pages={13142-13153}
}

@inproceedings{liu2023zero1to3,
      title={{Zero-1-to-3: Zero-shot One Image to 3D Object}}, 
      author={Ruoshi Liu and Rundi Wu and Basile Van Hoorick and Pavel Tokmakov and Sergey Zakharov and Carl Vondrick},
      year={2023},
      booktitle=ICCV
}

@inproceedings{Wen2023FoundationPoseU6,
  title={{FoundationPose: Unified 6D Pose Estimation and Tracking of Novel Objects}},
  author={Bowen Wen and Wei Yang and Jan Kautz and Stanley T. Birchfield},
  booktitle = CVPR,
  year={2023},
  pages={17868-17879}
}

@misc{shi2025nvspolicyadaptivenovelviewsynthesis,
      title={{NVSPolicy: Adaptive Novel-View Synthesis for Generalizable Language-Conditioned Policy Learning}}, 
      author={Le Shi and Yifei Shi and Xin Xu and Tenglong Liu and Junhua Xi and Chengyuan Chen},
      year={2025},
      eprint={2505.10359},
      archivePrefix={arXiv},
      primaryClass={cs.RO},
      url={https://arxiv.org/abs/2505.10359}, 
}

@inproceedings{Peng2018PVNetPV,
  title={PVNet: Pixel-Wise Voting Network for 6DoF Pose Estimation},
  author={Sida Peng and Yuan Liu and Qi-Xing Huang and Hujun Bao and Xiaowei Zhou},
  booktitle = CVPR,
  year={2018},
  pages={4556-4565}
}

@inproceedings{zhang2025recapture,
    title={ReCapture: Generative Video Camera Controls for User-Provided Videos using Masked Video Fine-Tuning},
    author={Zhang, David Junhao and Paiss, Roni and Zada, Shiran and Karnad, Nikhil and Jacobs, David E and Pritch, Yael and Mosseri, Inbar and Shou, Mike Zheng and Wadhwa, Neal and Ruiz, Nataniel},
    booktitle = CVPR,
    year={2025}
}

@article{lhhuang2024iclora,
  title={{In-Context LoRA for Diffusion Transformers}},
  author={Huang, Lianghua and Wang, Wei and Wu, Zhi-Fan and Shi, Yupeng and Dou, Huanzhang and Liang, Chen and Feng, Yutong and Liu, Yu and Zhou, Jingren},
  journal={arXiv preprint arxiv:2410.23775},
  year={2024}
}

@inproceedings{zhou2025stable,
              title={{Stable Virtual Camera: Generative View Synthesis with Diffusion Models}},
              author={Jensen (Jinghao) Zhou and Hang Gao and Vikram Voleti and Aaryaman Vasishta and Chun-Han Yao and Mark Boss and
              Philip Torr and Christian Rupprecht and Varun Jampani
              },
              booktitle = ICCV,
              year={2025}
          }

@inproceedings{
liu2024syncdreamer,
title={{SyncDreamer: Generating Multiview-consistent Images from a Single-view Image}},
author={Yuan Liu and Cheng Lin and Zijiao Zeng and Xiaoxiao Long and Lingjie Liu and Taku Komura and Wenping Wang},
booktitle= ICLR,
year={2024},
url={https://openreview.net/forum?id=MN3yH2ovHb}
}

@InProceedings{yu2025trajectory,
    author    = {Yu, Mark and Hu, Wenbo and Xing, Jinbo and Shan, Ying},
    title     = {{TrajectoryCrafter: Redirecting Camera Trajectory for Monocular Videos via Diffusion Models}},
    booktitle = ICCV,
    year      = {2025}
}

@inproceedings{Szymanowicz2023ViewsetD,
  title={{Viewset Diffusion: (0-)Image-Conditioned 3D Generative Models from 2D Data}},
  author={Stanislaw Szymanowicz and C. Rupprecht and Andrea Vedaldi},
  booktitle=ICCV,
  year={2023},
  pages={8829-8839}
}

@inproceedings{Sun2022OnePoseOO,
  title={{OnePose: One-Shot Object Pose Estimation without CAD Models}},
  author={Jiaming Sun and Zihao Wang and Siyu Zhang and Xingyi He He and Hongcheng Zhao and Guofeng Zhang and Xiaowei Zhou},
  booktitle = CVPR,
  year={2022},
  pages={6815-6824}
}

@misc{shi2023zero123plus,
      title={{Zero123++: a Single Image to Consistent Multi-view Diffusion Base Model}}, 
      author={Ruoxi Shi and Hansheng Chen and Zhuoyang Zhang and Minghua Liu and Chao Xu and Xinyue Wei and Linghao Chen and Chong Zeng and Hao Su},
      year={2023},
      eprint={2310.15110},
      archivePrefix={arXiv},
      primaryClass={cs.CV}
}

@inproceedings{ren2025gen3c,
        title={{GEN3C: 3D-Informed World-Consistent Video Generation with Precise Camera Control}},
        author={Ren, Xuanchi and Shen, Tianchang and Huang, Jiahui and Ling, Huan and 
            Lu, Yifan and Nimier-David, Merlin and M\"uller, Thomas and Keller, Alexander and 
            Fidler, Sanja and Gao, Jun},
        booktitle=CVPR,
        year={2025}
    }

@inproceedings{stojanov21toys4k,
      title={{Using Shape to Categorize: Low-Shot Learning with an Explicit Shape Bias}},
      author={Stefan Stojanov and Anh Thai and James M. Rehg},
      booktitle = CVPR,
      year      = {2021}
}

@article{li2025droplet3d,
      title={{Droplet3D: Commonsense Priors from Videos Facilitate 3D Generation}},
      author={Li, Xiaochuan and Du, Guoguang and Zhang, Runze and Jin, Liang and Jia, Qi and Lu, Lihua and Guo, Zhenhua and Zhao, Yaqian and Liu, Haiyang and Wang, Tianqi and Li, Changsheng and Gong, Xiaoli and Li, Rengang and Fan, Baoyu},
      journal={arXiv preprint arXiv:2508.20470},
      year={2025}
}

@inproceedings{voleti2024sv3d,
  author    = {Voleti, Vikram and Yao, Chun-Han and Boss, Mark and Letts, Adam and Pankratz, David and Tochilkin,  Dmitrii and Laforte, Christian and Rombach, Robin and Jampani, Varun},
  title     = {{SV3D: Novel Multi-view Synthesis and 3D Generation from a Single Image using Latent Video Diffusion}},
  booktitle = ECCV,
  year      = {2024},
}

@article{liu2023one2345,
  title={{One-2-3-45: Any single image to 3d mesh in 45 seconds without per-shape optimization}},
  author={Liu, Minghua and Xu, Chao and Jin, Haian and Chen, Linghao and Varma T, Mukund and Xu, Zexiang and Su, Hao},
  journal=NeurIPS,
  volume={36},
  year={2024}
}

@inproceedings{liu2023one2345++,
  title={{One-2-3-45++: Fast Single Image to 3D Objects with Consistent Multi-View Generation and 3D Diffusion}},
  author={Minghua Liu and Ruoxi Shi and Linghao Chen and Zhuoyang Zhang and Chao Xu and Xinyue Wei and Hansheng Chen and Chong Zeng and Jiayuan Gu and Hao Su},
  booktitle = CVPR,
  year={2024},
  url={https://arxiv.org/pdf/2311.07885.pdf}
}

@inproceedings{xu2024grm,
  title={{GRM: Large Gaussian Reconstruction Model for
Efficient 3D Reconstruction and Generation}},
  author={Xu, Yinghao and Shi, Zifan and Yifan, Wang and Chen, Hansheng and Yang, Ceyuan and Peng, Sida and Shen, Yujun and Wetzstein, Gordon},
  booktitle=ECCV,
  year={2024}
}

@inproceedings{tseng2023consistent,
  title={{Consistent view synthesis with pose-guided diffusion models}},
  author={Tseng, Hung-Yu and Li, Qinbo and Kim, Changil and Alsisan, Suhib and Huang, Jia-Bin and Kopf, Johannes},
  booktitle=CVPR,
  year={2023},
  url={https://arxiv.org/abs/2303.17598}
}

@inproceedings{
watson2023novel,
title={{Novel View Synthesis with Diffusion Models}},
author={Daniel Watson and William Chan and Ricardo Martin Brualla and Jonathan Ho and Andrea Tagliasacchi and Mohammad Norouzi},
booktitle=ICLR,
year={2023},
url={https://openreview.net/forum?id=HtoA0oT30jC}
}

@article{Deitke2023ObjaverseXLAU,
  title={{Objaverse-XL: A Universe of 10M+ 3D Objects}},
  author={Matt Deitke and Ruoshi Liu and Matthew Wallingford and Huong Ngo and Oscar Michel and Aditya Kusupati and Alan Fan and Christian Laforte and Vikram S. Voleti and Samir Yitzhak Gadre and Eli VanderBilt and Aniruddha Kembhavi and Carl Vondrick and Georgia Gkioxari and Kiana Ehsani and Ludwig Schmidt and Ali Farhadi},
  journal=NeurIPS,
  volume={36},
  pages={35799--35813},
  year={2023}
}

@misc{hunyuan3d2025hunyuan3d,
    title={{Hunyuan3D 2.1: From Images to High-Fidelity 3D Assets with Production-Ready PBR Material}},
    author={Tencent Hunyuan3D Team},
    year={2025},
    eprint={2506.15442},
    archivePrefix={arXiv},
    primaryClass={cs.CV}
}

@article{Downs2022GoogleSO,
  title={{Google Scanned Objects: A High-Quality Dataset of 3D Scanned Household Items}},
  author={Laura Downs and Anthony Francis and Nate Koenig and Brandon Kinman and Ryan Michael Hickman and Krista Reymann and Thomas Barlow McHugh and Vincent Vanhoucke},
  journal={International Conference on Robotics and Automation (ICRA)},
  year={2022},
  pages={2553-2560}
}

@inproceedings{long2024wonder3d,
  title={{Wonder3D: Single Image to 3D using Cross-Domain Diffusion}},
  author={Long, Xiaoxiao and Guo, Yuan-Chen and Lin, Cheng and Liu, Yuan and Dou, Zhiyang and Liu, Lingjie and Ma, Yuexin and Zhang, Song-Hai and Habermann, Marc and Theobalt, Christian and others},
  booktitle = CVPR,
  year={2024}
}

@inproceedings{
bahmani2025vdd,
title={{VD}3D: Taming Large Video Diffusion Transformers for 3D Camera Control},
author={Sherwin Bahmani and Ivan Skorokhodov and Aliaksandr Siarohin and Willi Menapace and Guocheng Qian and Michael Vasilkovsky and Hsin-Ying Lee and Chaoyang Wang and Jiaxu Zou and Andrea Tagliasacchi and David B. Lindell and Sergey Tulyakov},
booktitle=ICLR,
year={2025},
url={https://openreview.net/forum?id=0n4bS0R5MM}
}

@misc{wu2025qwenimagetechnicalreport,
      title={Qwen-Image Technical Report}, 
      author={Chenfei Wu and Jiahao Li and Jingren Zhou and Junyang Lin and Kaiyuan Gao and Kun Yan and Sheng-ming Yin and Shuai Bai and Xiao Xu and Yilei Chen and Yuxiang Chen and Zecheng Tang and Zekai Zhang and Zhengyi Wang and An Yang and Bowen Yu and Chen Cheng and Dayiheng Liu and Deqing Li and Hang Zhang and Hao Meng and Hu Wei and Jingyuan Ni and Kai Chen and Kuan Cao and Liang Peng and Lin Qu and Minggang Wu and Peng Wang and Shuting Yu and Tingkun Wen and Wensen Feng and Xiaoxiao Xu and Yi Wang and Yichang Zhang and Yongqiang Zhu and Yujia Wu and Yuxuan Cai and Zenan Liu},
      year={2025},
      eprint={2508.02324},
      archivePrefix={arXiv},
      primaryClass={cs.CV},
      url={https://arxiv.org/abs/2508.02324}, 
}

@inproceedings{wang2019normalized,
  title={Normalized object coordinate space for category-level 6d object pose and size estimation},
  author={Wang, He and Sridhar, Srinath and Huang, Jingwei and Valentin, Julien and Song, Shuran and Guibas, Leonidas J},
  booktitle=CVPR,
  pages={2642--2651},
  year={2019}
}

@misc{
huang2025cupid,
title={{CUPID}: Pose-Grounded Generative 3D Reconstruction from a Single Image},
author={Binbin Huang and Haobin Duan and Yiqun Zhao and Zibo Zhao and Yi Ma and Shenghua Gao},
year={2025},
url={https://openreview.net/forum?id=qSCjC9PFhs}
}

@misc{zhang2025easycontroladdingefficientflexible,
      title={{EasyControl: Adding Efficient and Flexible Control for Diffusion Transformer}}, 
      author={Yuxuan Zhang and Yirui Yuan and Yiren Song and Haofan Wang and Jiaming Liu},
      year={2025},
      eprint={2503.07027},
      archivePrefix={arXiv},
      primaryClass={cs.CV},
      url={https://arxiv.org/abs/2503.07027}, 
}

@inproceedings{wang2024motionctrl,
  title={MotionCtrl: A Unified and Flexible Motion Controller for Video Generation},
  author={Wang, Zhouxia and Yuan, Ziyang and Wang, Xintao and Li, Yaowei and Chen, Tianshui and Xia, Menghan and Luo, Ping and Shan, Ying},
  booktitle={ACM SIGGRAPH 2024 Conference Papers},
  year={2023}
}

@inproceedings{chen2020category,
  title={Category level object pose estimation via neural analysis-by-synthesis},
  author={Chen, Xu and Dong, Zijian and Song, Jie and Geiger, Andreas and Hilliges, Otmar},
  booktitle=ECCV,
  year={2020}
}

@inproceedings{
wang2025orient,
title={{Orient Anything: Learning Robust Object Orientation Estimation from Rendering 3D Models}},
author={Zehan Wang and Ziang Zhang and Tianyu Pang and Chao Du and Hengshuang Zhao and Zhou Zhao},
booktitle=ICML,
year={2025},
url={https://openreview.net/forum?id=x4yTgv2WkJ}
}

@inproceedings{wangorientv2,
title={Orient Anything V2: Unifying Orientation and Rotation Understanding},
author={Wang, Zehan and Zhang, Ziang and Xu, Jiayang and Wang, Jialei and Pang, Tianyu and Du, Chao and Zhao, Hengshuang and Zhao, Zhou},
booktitle={The Thirty-ninth Annual Conference on Neural Information Processing Systems},year    = {2026},
}

@inproceedings{xiang2024structured,
    title   = {{Structured 3D Latents for Scalable and Versatile 3D Generation}},
    author  = {Xiang, Jianfeng and Lv, Zelong and Xu, Sicheng and Deng, Yu and Wang, Ruicheng and 
               Zhang, Bowen and Chen, Dong and Tong, Xin and Yang, Jiaolong},
    booktitle = CVPR,
    year    = {2025},
    url={https://arxiv.org/abs/2412.01506}
}

@inproceedings{
hu2022lora,
title={Lo{RA}: Low-Rank Adaptation of Large Language Models},
author={Edward J Hu and Yelong Shen and Phillip Wallis and Zeyuan Allen-Zhu and Yuanzhi Li and Shean Wang and Lu Wang and Weizhu Chen},
booktitle={International Conference on Learning Representations},
year={2022},
url={https://openreview.net/forum?id=nZeVKeeFYf9}
}

@misc{
      lu2025orientation,
      title={Orientation Matters: Making 3D Generative Models Orientation-Aligned}, 
      author={Yichong Lu and Yuzhuo Tian and Zijin Jiang and Yikun Zhao and Yuanbo Yang and Hao Ouyang and Haoji Hu and Huimin Yu and Yujun Shen and Yiyi Liao},
      year={2025},
      eprint={2506.08640},
      archivePrefix={arXiv},
      primaryClass={cs.CV},
      url={https://arxiv.org/abs/2506.08640}, 
}

@misc{claude45,
  author       = {Anthropic},
  title        = {Claude Opus 4.5},
  year         = {2025},
  howpublished = {Anthropic News},
  url          = {https://www.anthropic.com/news/claude-sonnet-4-5},
}

@article{kong2024eschernet,
    title={EscherNet: A Generative Model for Scalable View Synthesis},
  author={Kong, Xin and Liu, Shikun and Lyu, Xiaoyang and Taher, Marwan and Qi, Xiaojuan and Davison, Andrew J},
  journal={arXiv preprint arXiv:2402.03908},
  year={2024}
}

@article{collins2022abo,
  title={ABO: Dataset and Benchmarks for Real-World 3D Object Understanding},
  author={Collins, Jasmine and Goel, Shubham and Deng, Kenan and Luthra, Achleshwar and
          Xu, Leon and Gundogdu, Erhan and Zhang, Xi and Yago Vicente, Tomas F and
          Dideriksen, Thomas and Arora, Himanshu and Guillaumin, Matthieu and
          Malik, Jitendra},
  journal={CVPR},
  year={2022}
}

@misc{feng2025seed3d10imageshighfidelity,
      title={Seed3D 1.0: From Images to High-Fidelity Simulation-Ready 3D Assets}, 
      author={Jiashi Feng and Xiu Li and Jing Lin and Jiahang Liu and Gaohong Liu and Weiqiang Lou and Su Ma and Guang Shi and Qinlong Wang and Jun Wang and Zhongcong Xu and Xuanyu Yi and Zihao Yu and Jianfeng Zhang and Yifan Zhu and Rui Chen and Jinxin Chi and Zixian Du and Li Han and Lixin Huang and Kaihua Jiang and Yuhan Li and Guan Luo and Shuguang Wang and Qianyi Wu and Fan Yang and Junyang Zhang and Xuanmeng Zhang},
      year={2025},
      eprint={2510.19944},
      archivePrefix={arXiv},
      primaryClass={eess.IV},
      url={https://arxiv.org/abs/2510.19944}, 
}

@article{liang2025UnitTEX,
  title={UniTEX: Universal High Fidelity Generative Texturing for 3D Shapes},
  author={Yixun Liang and Kunming Luo and Xiao Chen and Rui Chen and Hongyu Yan and Weiyu Li and Jiarui Liu and Ping Tan},
  journal={arXiv preprint arXiv:2505.23253},
  year={2025}
}

@misc{hunyuan3d2026hy3dbenchgeneration3dassets,
      title={HY3D-Bench: Generation of 3D Assets}, 
      author={Team Hunyuan3D and : and Bowen Zhang and Chunchao Guo and Dongyuan Guo and Haolin Liu and Hongyu Yan and Huiwen Shi and Jiaao Yu and Jiachen Xu and Jingwei Huang and Kunhong Li and Lifu Wang and Linus and Penghao Wang and Qingxiang Lin and Ruining Tang and Xianghui Yang and Yang Li and Yirui Guan and Yunfei Zhao and Yunhan Yang and Zeqiang Lai and Zhihao Liang and Zibo Zhao},
      year={2026},
      eprint={2602.03907},
      archivePrefix={arXiv},
      primaryClass={cs.CV},
      url={https://arxiv.org/abs/2602.03907}, 
}

@inproceedings{gortler1996lumigraph,
  title={The lumigraph},
  author={Gortler, Steven J and Grzeszczuk, Radek and Szeliski, Richard and Cohen, Michael F},
  booktitle={Proceedings of the 23rd annual conference on Computer graphics and interactive techniques},
  pages={43--54},
  year={1996}
}

@inproceedings{levoy1996light,
  title={Light field rendering},
  author={Levoy, Marc and Hanrahan, Pat},
  booktitle={Proceedings of the 23rd annual conference on Computer graphics and interactive techniques},
  pages={31--42},
  year={1996}
}

@inproceedings{mildenhall2020nerf,
  title={NeRF: Representing Scenes as Neural Radiance Fields for View Synthesis},
  author={Mildenhall, Ben and Srinivasan, Pratul P and Tancik, Matthew and Barron, Jonathan T and Ramamoorthi, Ravi and Ng, Ren},
  booktitle={European Conference on Computer Vision},
  pages={405--421},
  year={2020},
  organization={Springer}
}

@article{muller2022instant,
  title={Instant neural graphics primitives with a multiresolution hash encoding},
  author={M{\"u}ller, Thomas and Evans, Alex and Schied, Christoph and Keller, Alexander},
  journal={ACM Transactions on Graphics (ToG)},
  volume={41},
  number={4},
  pages={1--15},
  year={2022},
  publisher={ACM New York, NY, USA}
}

@inproceedings{yu2021pixelnerf,
  title={pixelNeRF: Neural radiance fields from one or few images},
  author={Yu, Alex and Ye, Vickie and Tancik, Matthew and Kanazawa, Angjoo},
  booktitle={Proceedings of the IEEE/CVF Conference on Computer Vision and Pattern Recognition},
  pages={4578--4587},
  year={2021}
}

@article{kerbl20233d,
  title={3D Gaussian Splatting for Real-Time Radiance Field Rendering},
  author={Kerbl, Bernhard and Kopanas, Georgios and Leimk{\"u}hler, Thomas and Drettakis, George},
  journal={ACM Transactions on Graphics (ToG)},
  volume={42},
  number={4},
  pages={1--14},
  year={2023},
  publisher={ACM New York, NY, USA}
}

@inproceedings{szymanowicz2024splatter,
  title={Splatter image: Ultra-fast single-view 3d reconstruction},
  author={Szymanowicz, Stanislaw and Rupprecht, Christian and Vedaldi, Andrea},
  booktitle={Proceedings of the IEEE/CVF Conference on Computer Vision and Pattern Recognition},
  pages={20892--20901},
  year={2024}
}

@inproceedings{charatan2024pixelsplat,
  title={pixelsplat: 3d Gaussian Splats from Image Pairs for Scalable Generalizable 3D Reconstruction},
  author={Charatan, David and Li, Sizhe and Tagliasacchi, Andrea and Sitzmann, Vincent},
  booktitle={Proceedings of the IEEE/CVF Conference on Computer Vision and Pattern Recognition},
  pages={20912--20922},
  year={2024}
}

@inproceedings{chen2024mvsplat,
  title={MVSplat: Efficient 3D Gaussian Splatting from Sparse Multi-view Images},
  author={Chen, Yuedong and Xu, Haofei and Zheng, Chuanxia and Zhuang, Bohan and Pollefeys, Marc and Geiger, Andreas and Cham, Tat-Jen and Cai, Jianfei},
  booktitle={European Conference on Computer Vision},
  year={2024},
  organization={Springer}
}

@inproceedings{ho2020denoising,
  title={Denoising diffusion probabilistic models},
  author={Ho, Jonathan and Jain, Ajay and Abbeel, Pieter},
  booktitle={Advances in Neural Information Processing Systems},
  volume={33},
  pages={6840--6851},
  year={2020}
}

@inproceedings{song2019generative,
  title={Generative modeling by estimating gradients of the data distribution},
  author={Song, Yang and Ermon, Stefano},
  booktitle={Advances in Neural Information Processing Systems},
  volume={32},
  year={2019}
}

@inproceedings{wang2023score,
  title={Score jacobian chaining: Lifting pretrained 2d diffusion models for 3d generation},
  author={Wang, Haochen and Du, Xiaodan and Li, Jiahao and Yeh, Raymond A and Shakhnarovich, Greg},
  booktitle={Proceedings of the IEEE/CVF Conference on Computer Vision and Pattern Recognition},
  pages={12619--12629},
  year={2023}
}

@inproceedings{poole2023dreamfusion,
  title={DreamFusion: Text-to-3D using 2D Diffusion},
  author={Poole, Ben and Jain, Ajay and Barron, Jonathan T and Mildenhall, Ben},
  booktitle={International Conference on Learning Representations},
  year={2023}
}

@inproceedings{wu2024reconfusion,
  title={ReconFusion: 3d Reconstruction with Diffusion Priors},
  author={Wu, Rundi and Mildenhall, Ben and Henzler, Philipp and Park, Keunhong and Gao, Ruiqi and Watson, Daniel and Srinivasan, Pratul P and Verbin, Dor and Barron, Jonathan T and Poole, Ben and others},
  booktitle={Proceedings of the IEEE/CVF Conference on Computer Vision and Pattern Recognition},
  pages={21382--21392},
  year={2024}
}

@article{wang2023imagedream,
  title={ImageDream: Image-Prompt Multi-view Diffusion for 3D Generation},
  author={Wang, Peng and Shi, Yichun},
  journal={arXiv preprint arXiv:2312.02201},
  year={2023}
}

@article{gao2024cat3d,
    title={CAT3D: Create Anything in 3D with Multi-View Diffusion Models},
    author={Ruiqi Gao* and Aleksander Holynski* and Philipp Henzler and Arthur Brussee and Ricardo Martin-Brualla and Pratul P. Srinivasan and Jonathan T. Barron and Ben Poole*
    },
    journal={Advances in Neural Information Processing Systems},
    year={2024}
}

@inproceedings{sajjadi2022scene,
  title={Scene Representation Transformer: Geometry-Free Novel View Synthesis Through Set-Latent Scene Representations},
  author={Mehdi S. M. Sajjadi and Henning Meyer and Etienne Pot and Urs Bergmann and Klaus Greff and Noha Radwan and Suhani Vora and Mario Lučić and Daniel Duckworth and Alexey Dosovitskiy and Jakob Uszkoreit and Tom Funkhouser and Andrea Tagliasacchi},
  booktitle={Proceedings of the IEEE/CVF Conference on Computer Vision and Pattern Recognition},
  pages={16292--16301},
  year={2022}
}

@inproceedings{jin2025lvsm,
  title={LVSM: A Large View Synthesis Model with Minimal 3D Inductive Bias},
  author={Jin, Haian and Jiang, Hanwen and Tan, Hao and Zhang, Kai and Bi, Sai and Zhang, Tianyuan and Luan, Fujun and Snavely, Noah and Xu, Zexiang},
  booktitle={International Conference on Learning Representations},
  year={2025}
}

@inproceedings{liu2023zero,
  title={Zero-1-to-3: Zero-shot one image to 3d object},
  author={Liu, Ruoshi and Wu, Rundi and Van Hoorick, Basile and Tokmakov, Pavel and Zakharov, Sergey and Vondrick, Carl},
  booktitle={Proceedings of the IEEE/CVF International Conference on Computer Vision},
  pages={9298--9309},
  year={2023}
}

@misc{he2024cameractrl,
    title={CameraCtrl: Enabling Camera Control for Text-to-Video Generation},
    author={Hao He and Yinghao Xu and Yuwei Guo and Gordon Wetzstein and Bo Dai and Hongsheng Li and Ceyuan Yang},
    year={2024},
    eprint={2404.02101},
    archivePrefix={arXiv},
    primaryClass={cs.CV}
}

@inproceedings{van2024generative,
  title={Generative Camera Dolly: Extreme Monocular Dynamic Novel View Synthesis},
  author={Van Hoorick, Basile and Wu, Rundi and Ozguroglu, Ege and Sargent, Kyle and Liu, Ruoshi and Tokmakov, Pavel and Dave, Achal and Zheng, Changxi and Vondrick, Carl},
  booktitle={European Conference on Computer Vision},
  year={2024},
  organization={Springer}
}

@inproceedings{hong2024lrm,
  title={Lrm: Large reconstruction model for single image to 3d},
  author={Hong, Yicong and Zhang, Kai and Gu, Jiuxiang and Bi, Sai and Yang, Yang and Forsyth, David and others},
  booktitle={International Conference on Learning Representations},
  year={2024}
}

@inproceedings{goodfellow2014generative,
  title={Generative Adversarial Networks},
  author={Goodfellow, Ian and Pouget-Abadie, Jean and Mirza, Mehdi and Xu, Bing and Warde-Farley, David and Ozair, Sherjil and Courville, Aaron and Bengio, Yoshua},
  booktitle={Advances in neural information processing systems},
  volume={27},
  pages={2672--2680},
  year={2014}
}

@inproceedings{wu2016learning,
  title     = {Learning a Probabilistic Latent Space of Object Shapes via 3D Generative-Adversarial Modeling},
  author    = {Wu, Jiajun and Zhang, Chengkai and Xue, Tianfan and Freeman, William T and Tenenbaum, Joshua B},
  booktitle = {Advances in Neural Information Processing Systems (NeurIPS)},
  volume    = {29},
  pages     = {82--90},
  year      = {2016}
}

@inproceedings{Chan2021,
            author = {Eric R. Chan and Connor Z. Lin and Matthew A. Chan and Koki Nagano and Boxiao Pan and Shalini De Mello and Orazio Gallo and Leonidas Guibas and Jonathan Tremblay and Sameh Khamis and Tero Karras and Gordon Wetzstein},
            title = {Efficient Geometry-aware {3D} Generative Adversarial Networks},
            booktitle = {arXiv},
            year = {2021}
          }

@article{rafailov2023direct,
  title={Direct preference optimization: Your language model is secretly a reward model},
  author={Rafailov, Rafael and Sharma, Archit and Mitchell, Eric and Manning, Christopher D and Ermon, Stefano and Finn, Chelsea},
  journal={Advances in neural information processing systems},
  volume={36},
  pages={53728--53741},
  year={2023}
}

@article{shao2024deepseekmath,
  title={Deepseekmath: Pushing the limits of mathematical reasoning in open language models},
  author={Shao, Zhihong and Wang, Peiyi and Zhu, Qihao and Xu, Runxin and Song, Junxiao and Bi, Xiao and Zhang, Haowei and Zhang, Mingchuan and Li, YK and Wu, Yang and others},
  journal={arXiv preprint arXiv:2402.03300},
  year={2024}
}

@misc{bai2025qwen25vltechnicalreport,
      title={Qwen2.5-VL Technical Report}, 
      author={Shuai Bai and Keqin Chen and Xuejing Liu and Jialin Wang and Wenbin Ge and Sibo Song and Kai Dang and Peng Wang and Shijie Wang and Jun Tang and Humen Zhong and Yuanzhi Zhu and Mingkun Yang and Zhaohai Li and Jianqiang Wan and Pengfei Wang and Wei Ding and Zheren Fu and Yiheng Xu and Jiabo Ye and Xi Zhang and Tianbao Xie and Zesen Cheng and Hang Zhang and Zhibo Yang and Haiyang Xu and Junyang Lin},
      year={2025},
      eprint={2502.13923},
      archivePrefix={arXiv},
      primaryClass={cs.CV},
      url={https://arxiv.org/abs/2502.13923}, 
}

@article{li2025cameras,
  title={Cameras as relative positional encoding},
  author={Li, Ruilong and Yi, Brent and Liu, Junchen and Gao, Hang and Ma, Yi and Kanazawa, Angjoo},
  journal={arXiv preprint arXiv:2507.10496},
  year={2025}
}

@article{miyato2023gta,
  title={Gta: A geometry-aware attention mechanism for multi-view transformers},
  author={Miyato, Takeru and Jaeger, Bernhard and Welling, Max and Geiger, Andreas},
  journal={arXiv preprint arXiv:2310.10375},
  year={2023}
}

@inproceedings{liu2026kaleido,
  title={Scaling Sequence-to-Sequence Generative Neural Rendering},
  author={Liu, Shikun and Ng, Kam Woh and Jang, Wonbong and Guo, Jiadong and Han, Junlin and Liu, Haozhe and Douratsos, Yiannis and P{\'e}rez, Juan C and Zhou, Zijian and Phung, Chi and others},
  booktitle={International Conference on Learning Representations ({ICLR})},
  year={2026}
}

@article{wiedemer2025video,
  title={Video models are zero-shot learners and reasoners},
  author={Wiedemer, Thadd{\"a}us and Li, Yuxuan and Vicol, Paul and Gu, Shixiang Shane and Matarese, Nick and Swersky, Kevin and Kim, Been and Jaini, Priyank and Geirhos, Robert},
  journal={arXiv preprint arXiv:2509.20328},
  year={2025}
}

@article{wei2022chain,
  title={Chain-of-thought prompting elicits reasoning in large language models},
  author={Wei, Jason and Wang, Xuezhi and Schuurmans, Dale and Bosma, Maarten and Xia, Fei and Chi, Ed and Le, Quoc V and Zhou, Denny and others},
  journal={Advances in neural information processing systems},
  volume={35},
  pages={24824--24837},
  year={2022}
}

@article{
    xiang2025trellis2,
    title={Native and Compact Structured Latents for 3D Generation},
    author={Xiang, Jianfeng and Chen, Xiaoxue and Xu, Sicheng and Wang, Ruicheng and Lv, Zelong and Deng, Yu and Zhu, Hongyuan and Dong, Yue and Zhao, Hao and Yuan, Nicholas Jing and Yang, Jiaolong},
    journal={Tech report},
    year={2025}
}

@inproceedings{esser2024scaling,
  title={Scaling rectified flow transformers for high-resolution image synthesis},
  author={Esser, Patrick and Kulal, Sumith and Blattmann, Andreas and Entezari, Rahim and M{\"u}ller, Jonas and Saini, Harry and Levi, Yam and Lorenz, Dominik and Sauer, Axel and Boesel, Frederic and others},
  booktitle={Forty-first international conference on machine learning},
  year={2024}
}

@inproceedings{zhang2018unreasonable,
  title={The unreasonable effectiveness of deep features as a perceptual metric},
  author={Zhang, Richard and Isola, Phillip and Efros, Alexei A and Shechtman, Eli and Wang, Oliver},
  booktitle={Proceedings of the IEEE conference on computer vision and pattern recognition},
  pages={586--595},
  year={2018}
}

@inproceedings{radford2021learning,
  title={Learning transferable visual models from natural language supervision},
  author={Radford, Alec and Kim, Jong Wook and Hallacy, Chris and Ramesh, Aditya and Goh, Gabriel and Agarwal, Sandhini and Sastry, Girish and Askell, Amanda and Mishkin, Pamela and Clark, Jack and others},
  booktitle={International conference on machine learning},
  pages={8748--8763},
  year={2021},
  organization={PmLR}
}

@article{wan2025,
      title={Wan: Open and Advanced Large-Scale Video Generative Models}, 
      author={Team Wan and Ang Wang and Baole Ai and Bin Wen and Chaojie Mao and Chen-Wei Xie and Di Chen and Feiwu Yu and Haiming Zhao and Jianxiao Yang and Jianyuan Zeng and Jiayu Wang and Jingfeng Zhang and Jingren Zhou and Jinkai Wang and Jixuan Chen and Kai Zhu and Kang Zhao and Keyu Yan and Lianghua Huang and Mengyang Feng and Ningyi Zhang and Pandeng Li and Pingyu Wu and Ruihang Chu and Ruili Feng and Shiwei Zhang and Siyang Sun and Tao Fang and Tianxing Wang and Tianyi Gui and Tingyu Weng and Tong Shen and Wei Lin and Wei Wang and Wei Wang and Wenmeng Zhou and Wente Wang and Wenting Shen and Wenyuan Yu and Xianzhong Shi and Xiaoming Huang and Xin Xu and Yan Kou and Yangyu Lv and Yifei Li and Yijing Liu and Yiming Wang and Yingya Zhang and Yitong Huang and Yong Li and You Wu and Yu Liu and Yulin Pan and Yun Zheng and Yuntao Hong and Yupeng Shi and Yutong Feng and Zeyinzi Jiang and Zhen Han and Zhi-Fan Wu and Ziyu Liu},
      journal = {arXiv preprint arXiv:2503.20314},
      year={2025}
}

@article{yang2024cogvideox,
  title={CogVideoX: Text-to-Video Diffusion Models with An Expert Transformer},
  author={Yang, Zhuoyi and Teng, Jiayan and Zheng, Wendi and Ding, Ming and Huang, Shiyu and Xu, Jiazheng and Yang, Yuanming and Hong, Wenyi and Zhang, Xiaohan and Feng, Guanyu and others},
  journal={arXiv preprint arXiv:2408.06072},
  year={2024}
}

@article{melaskyriazi2024im3d,
    title={IM-3D: Iterative Multiview Diffusion and Reconstruction for High-Quality 3D Generation},
    author={Luke Melas-Kyriazi and Iro Laina and Christian Rupprecht and Natalia Neverova and Andrea Vedaldi and Oran Gafni and Filippos Kokkinos},
    journal={International Conference on Machine Learning, 2024},
    year={2024}
}

@article{bai2025recammaster,
  title={ReCamMaster: Camera-Controlled Generative Rendering from A Single Video},
  author={Bai, Jianhong and Xia, Menghan and Fu, Xiao and Wang, Xintao and Mu, Lianrui and Cao, Jinwen and Liu, Zuozhu and Hu, Haoji and Bai, Xiang and Wan, Pengfei and others},
  journal={arXiv preprint arXiv:2503.11647},
  year={2025}
}
\clearpage
\appendix
\setcounter{table}{0}
\setcounter{figure}{0}
\section{Preliminary: Qwen-Image-Edit}

For the sake of clarity, we provide a brief review of Qwen-Image-Edit~\cite{wu2025qwenimagetechnicalreport}, the state-of-the-art instruction-based image editing model that we build upon.

Qwen-Image-Edit is built on a Multi-Modal Diffusion Transformer (MMDiT)\cite{esser2024scaling} backbone that jointly models text and image tokens through shared transformer layers. The input image is processed via a dual-encoding strategy: a frozen Qwen2.5-VL\cite{bai2025qwen25vltechnicalreport} encoder extracts semantic features from its last hidden layer, while a joint image-video VAE encoder produces pixel-level latent representations. 
The model uses Multimodal Scalable RoPE (MSRoPE) for positional encoding, treating text as 2D tensors positioned along the diagonal of the image
grid, extending this with an additional \emph{frame} dimension to distinguish between input and output images tokens in editing tasks.

Trained with flow matching objectives and post-trained with DPO~\cite{rafailov2023direct} and GRPO~\cite{shao2024deepseekmath}, Qwen-Image-Edit can perform complex image manipulations off-the-shelf, including rudimentary novel view synthesis (NVS) through natural-language prompts such as \textit{``turn left/right 90 degrees''}. However, and unfortunately also common to other task-specific diffusion-based models for NVS~\cite{voleti2024sv3d,liu2023zero1to3}, Qwen-Image-Edit is unable to robustly perform \textit{global} viewpoint control operations due to the inherent complexity and ambiguity in defining a common reference frame. Additionally, as we show in our ablation study, text descriptions are inherently ambiguous and insufficient to specify exact camera parameters.

\newcommand{\PAR}[1]{\noindent\textbf{#1}}
\newcommand{\SUBPAR}[1]{\noindent\textit{\textbf{#1}}}

\section{Discussion: Choice of Base Model}
A natural question arises: \emph{why not adopt a video diffusion 
model as the base architecture for NVS?}
Video generation models~\cite{wan2025,yang2024cogvideox} maintain strong inter-frame consistency through built-in temporal modeling, and recent works have successfully repurposed them for multi-view generation~\cite{li2025droplet3d,voleti2024sv3d, melaskyriazi2024im3d} and camera-controlled generation~\cite{bai2025recammaster}. SEVA~\cite{zhou2025stable} directly fine-tunes for novel view synthesis. Despite these promising results, we identify two key reasons that make video models suboptimal for our setting.

\PAR{Image quality gap.}
Despite rapid progress, the visual fidelity of current video generation models still lags behind that of their image-generation counterparts at comparable resolution. This gap stems from the substantial computational overhead of modeling the temporal dimension, which in practice forces trade-offs in spatial resolution, sampling steps, or model capacity. Since our primary goal is to produce high-fidelity novel views, we opt to build upon a state-of-the-art image editing model whose output quality more closely matches that of frontier image generators.

\PAR{Architectural mismatch with NVS.}
More fundamentally, the architectural priors in video diffusion models are misaligned with the nature of the NVS task. Video models exploit the strong \emph{temporal continuity} of natural video through mechanisms such as temporal compression in the VAE latent space and temporal convolutions that capture smooth frame-to-frame transitions. These operations rely on a strict sequential ordering of frames—adjacent frames are expected to be visually similar, and the learned priors encode this locality. In NVS, however, the set of target views is \emph{fundamentally permutation-invariant}: the model should produce the same output regardless of the order in which target viewpoints are specified. Imposing a sequential temporal structure on an inherently unordered set of views introduces an unwanted inductive bias that can degrade both controllability and consistency, particularly when target views span large angular separations or non-sequential camera trajectories.

\PAR{Outlook.}
While the above considerations motivate our choice of an image editing backbone, we note that video data itself remains a valuable resource for NVS, as it naturally captures the spatial and geometric relationships between viewpoints. Incorporating video data during training or distilling complementary priors from video models into our framework represents a promising direction for future work.

\section{Implementation Details}

\subsection{Dataset Curation}

\PAR{Rendering.}
For each object, we render images at $1024 \times 1024$ resolution using Blender. We uniformly sample 34 viewpoints on a sphere surrounding the object, supplemented by 6 canonical views (front, back, left, right, top, and bottom), yielding 40 rendered images per object in total.

\PAR{VLM-Assisted Annotation.}
We employ a Vision-Language Model (VLM) for two annotation tasks: (1)~identifying the front-facing canonical view of each object, and (2)~generating structured captions comprising an object description and a front-view definition.

\SUBPAR{(1) Front-View Selection.}
Following the orientation conventions established by Orientation Matters~\cite{lu2025orientation}, we define category-specific rules for determining the front view of each object (\eg, the muzzle side for firearms, the head side for animals, the headlight/grille side for cars). Given four canonical views rendered in a fixed counter-clockwise order, we prompt the VLM to identify which image(s) correspond to the front of the object based on these semantic rules.

We observe that for elongated or thin objects (\eg, swords), the VLM struggles to distinguish between the front and back views, as these appear visually similar. However, the side views---where the object's elongated axis is clearly visible (\eg, the blade tip pointing left or right)---are far more discriminable. To exploit this, we adopt an indirect identification strategy: we first ask the VLM to identify the left-side view, and then determine the front view as the image immediately preceding it in the counter-clockwise rendering order. This approach significantly improves annotation accuracy for such ambiguous cases.

\SUBPAR{(2) Caption Generation.}
For each object, we prompt the VLM with four canonical views presented in a fixed order (front, right, back, left) and request two outputs:
\begin{itemize}
    \item \textit{Object description}: a view-invariant description of the object's identity, appearance, materials, colors, and distinguishing features, without any reference to camera viewpoints.
    \item \textit{Front-view definition}: a concise sentence characterizing the semantic signature of the object's front side (\eg, ``The front is defined by the headlights and grille'' for a car, or ``The front is defined by the dial face'' for a watch).
\end{itemize}
These captions serve as the text conditioning in our model, simultaneously disambiguating the canonical orientation and enabling generalization to unseen object categories at test time.

We provide the full prompts used for VLM-assisted annotation below.

\paragraph{Front-View Selection Prompt.}\mbox{}\\[-1.0\baselineskip]
\begin{lstlisting}[basicstyle=\ttfamily\scriptsize, breaklines=true, frame=single]
You are given 4 images of the same object, labeled A, B, C, D.
Each image shows the object from a different viewpoint 
(front / back / left / right).

Task: Decide which image(s) show the FRONT of the object.

General rule: The "front" is the direction the object 
naturally faces or moves toward.

Special rules (object-specific):
1) Weapons (gun, bow, etc.)
   - Front = muzzle / firing direction (or arrow direction).
2) Bicycle
   - Front = handlebar / front wheel direction.
3) Sofa
   - Front = side with the backrest.
4) Animals
   - Front = head direction.
5) Person
   - Front = face-facing side.
6) Car
   - Front = headlights / grille direction.
7) Watch
   - Front = dial face.
8) Teapot
   - Front = spout direction.
9) Cup / Mug
   - Front = the view where the handle points to the LEFT.
10) Sword (lying horizontal)
   - Side = blade tip on the left/right.
   - FRONT = the image immediately BEFORE the LEFT-side 
     view in the input order (A->B->C->D->A).

Other objects: Define front/back/side based on the 
object's structure and functional features.

Output format (JSON):
- Single front view: ["A"]
- Multiple front views: ["A", "B"]
\end{lstlisting}

\paragraph{Caption Generation Prompt.}\mbox{}\\[-1.0\baselineskip]
\begin{lstlisting}[basicstyle=\ttfamily\scriptsize, breaklines=true, frame=single]
You are labeling a single rigid object from 4 rendered views.
Input images are in this order:
(1) FRONT, (2) RIGHT, (3) BACK, (4) LEFT.

Your job:
1) "object_description": a view-invariant description of 
   the object's identity and appearance. Do NOT mention 
   camera/view words. Describe category, key parts, 
   materials, colors, patterns, logos/text. 2-3 sentences.

2) "object_front_in_world": one short sentence describing 
   the FRONT side's semantic indicator (e.g., face / 
   headlights / screen / spout / label side).
   Start with "The front is defined by".

Output format (JSON):
{
  "object_description": "...",
  "object_front_in_world": "..."
}
\end{lstlisting}

\subsection{Experiment Implementations}

We evaluate our method on two unseen datasets: GSO~\cite{Downs2022GoogleSO} and Toys4k~\cite{stojanov21toys4k}, selecting 100 objects per dataset across diverse categories, each with a clearly defined front view. All evaluations are conducted at $512 \times 512$ resolution.

For 3D generation methods without orientation awareness (Trellis-2~\cite{xiang2025trellis2} and Hunyuan3D 2.1~\cite{hunyuan3d2025hunyuan3d}), the generated meshes may be arbitrarily oriented. To ensure a fair comparison, we render each generated asset from four candidate front directions ($+x$, $-x$, $+y$, $-y$) and report the best metric score among them.
For 2D-based generative NVS methods that require input camera poses, we use Orient-Anything-V2~\cite{wangorientv2} to estimate the absolute object orientation (azimuth, polar, and roll) of each input view, which we then convert to the corresponding absolute camera pose in NOCS.

\section{Additional Results}

\subsection{Qualitative Results}

We provide further qualitative results, including real-world examples and application to 3D reconstruction. Please refer to \texttt{website/index.html} included in the supplementary material for full-resolution results.

\subsection{Baselines Using Ground-Truth Poses}

\begin{figure}[h]
    \centering
    \includegraphics[width=0.75\linewidth]{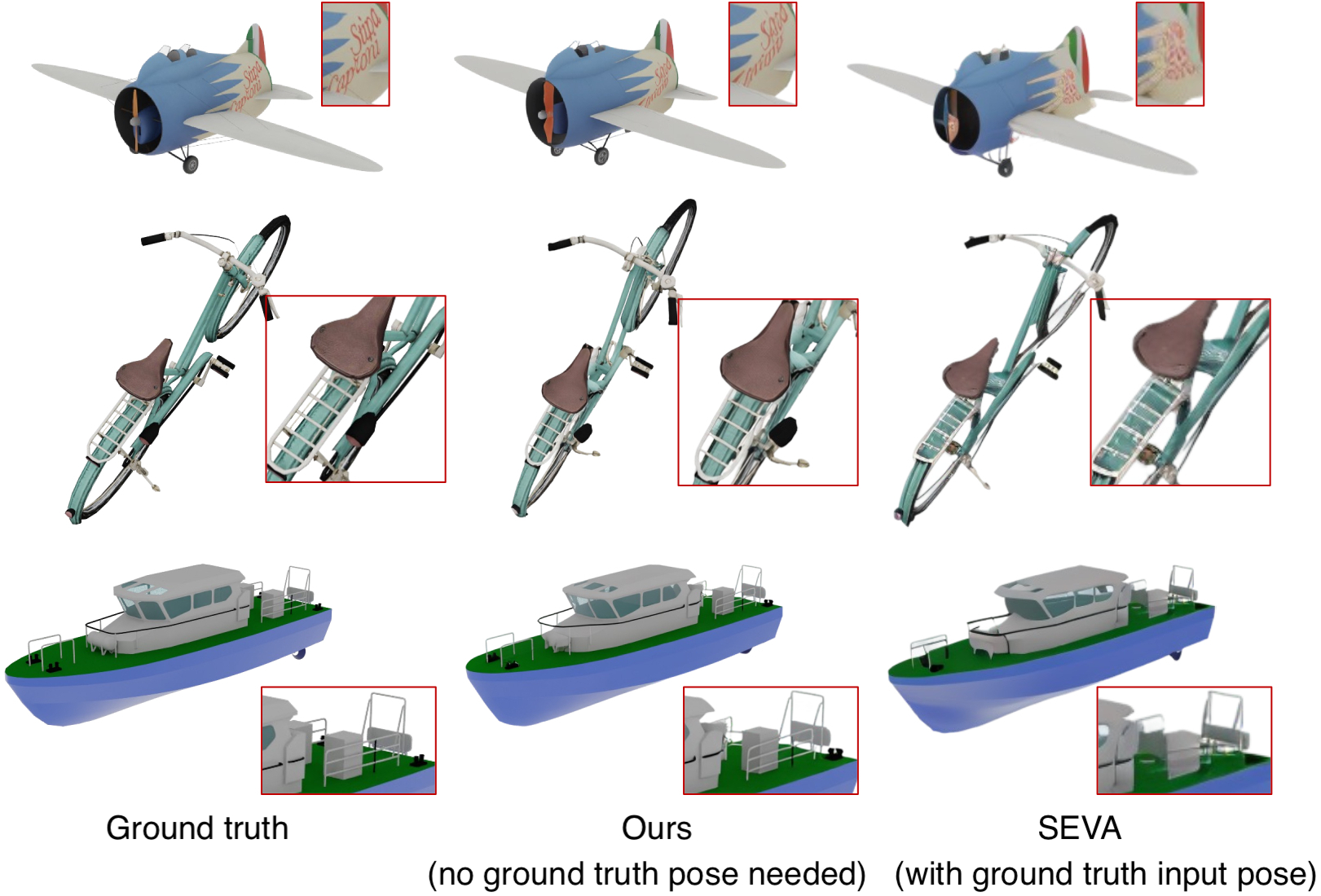}
    \caption{Ours (no pose input) \textit{vs.} SEVA (w/ GT input pose).}
    \label{fig:gtpose}
\end{figure}

As discussed in the main paper, we estimate camera poses for EscherNet~\cite{kong2024eschernet} and SEVA~\cite{zhou2025stable} to ensure evaluation under the same conditions, since these methods require camera poses as input while our approach does not.
Nonetheless, reporting results with ground-truth poses for pose-dependent baselines is informative.
Hence, we conduct an additional evaluation on Toys4k using 1, 2, and 3 views.
SEVA(GT) achieves PSNR values of 17.21, 20.17, and 21.35 dB, compared to 18.37, 18.93, and 19.31 dB for our approach.
While our method outperforms SEVA(GT) at 1 view, SEVA(GT) achieves higher PSNR with multiple views --- expected given its access to GT pose information, which improves pixel-level alignment --- though our method delivers visibly better overall generation quality (Figure~\ref{fig:gtpose}, \eg, structurally faithful bicycle frame).

We note that refining estimated poses would strengthen the baselines, but pose-dependent methods remain fundamentally limited to scenarios where camera poses can be reliably obtained. Our pose-free formulation directly addresses this limitation, making it a complementary capability. That said, extending our method to optionally accept camera poses is a promising direction for future work.

\subsection{Effect of Object Descriptions}

\begin{table}[h]
\centering
\caption{Caption ablation: \textit{Detailed} (full description) vs.\ \textit{Category-only} (class name only).}
\label{tab:comp_caption}
\begin{tabular}{llcccc}
\hline
\multicolumn{2}{c}{\#Input Views} & 1 & 2 & 3 & 5 \\
\hline
\multirow{3}{*}{Detailed}
& PSNR $\uparrow$ & 18.24 & 18.77 & 19.17 & 19.68 \\
& SSIM $\uparrow$ & 0.892 & 0.895 & 0.897 & 0.902 \\
& LPIPS $\downarrow$ & 0.116 & 0.107 & 0.100 & 0.093 \\
\hline
\multirow{3}{*}{Category-only}
& PSNR $\uparrow$ & 18.18 & 18.72 & 19.12 & 19.70 \\
& SSIM $\uparrow$ & 0.891 & 0.894 & 0.896 & 0.901 \\
& LPIPS $\downarrow$ & 0.117 & 0.108 & 0.101 & 0.093 \\
\hline
\end{tabular}
\end{table}

To assess the effect of the caption details, we ablate on Toys4k by replacing the detailed description (\textit{Detailed}) with only the object class name (\textit{Category-only}, \eg, \textit{airplane}, \textit{chair}), while retaining the front-face definition in both conditions.
As shown in Table~\ref{tab:comp_caption}, the \textit{Category-only} setting is highly competitive with \textit{Detailed}, and the gap further shrinks with more input views as visual evidence compensates for the weaker textual prior.
Regarding the source of captions: although Toys4k labels are dataset-provided, they can be reliably reproduced by a VLM, and our real-world results in the supplementary already use VLM-inferred captions.
We acknowledge that inaccurate captions may cause the model to hallucinate content, and we discuss this as a limitation.

\end{document}